\documentclass[a4paper]{article}
\usepackage[latin1]{inputenc}
\usepackage{latexsym}
\usepackage{amsmath}
\usepackage{amssymb}
\usepackage{mathptmx}
\usepackage{stmaryrd}

\usepackage{courier}
\usepackage{pstricks}
\usepackage{pifont}

\usepackage[amsmath,thmmarks,hyperref]{ntheorem}
\theoremstyle{plain} 
\theorembodyfont{\upshape} 
\theoremsymbol{\ensuremath{_\ast}}
\theoremseparator{\ } 
\newtheorem{theorem}{Theorem}
\newtheorem{proposition}{Proposition}
\newtheorem{lemma}{Lemma}
\newtheorem{definition}{Definition}
\newtheorem{example}{Example}

\theoremstyle{nonumberplain} 
\theoremheaderfont{\slshape} 
\theorembodyfont{\upshape} 
\theoremsymbol{\rule{.8ex}{.8ex}}
\theoremseparator{.} 
\newtheorem{proof}{Proof}

\newenvironment{mcqm@list}[1]{%
        \begin{list}{#1}{%
                \settowidth{\labelwidth}{\makelabel{\@itemlabel\hfill}}%
                \setlength\leftmargin\labelwidth
                \advance\leftmargin by\labelsep}}{%
        \end{list}}

\newenvironment{proofiff}{%
        \begin{mcqm@list}{%
                {${\Leftarrow}$}}%
                \newcommand*{\SUFF}{\item[{${\Leftarrow}$}]}%
                \newcommand*{\NECC}{\item[{${\Rightarrow}$}]}}{%
        \end{mcqm@list}}

\newcommand{\define}[1]{\emph{#1}}
\newcommand{\propref}[1]{\hyperref[#1]{P\ref{#1}}}

\usepackage{color}
\definecolor{gecodegreen}{rgb}{0.043,0.463,0.275}
\definecolor{gecodeblue}{rgb}{0.0,0.361,0.631}
\definecolor{gecodeorange}{rgb}{0.922,0.537,0.106}
\definecolor{gecodered}{rgb}{0.855,0.145,0.114}

\usepackage[
  colorlinks,
  linkcolor=gecodeblue,
  anchorcolor=gecodeblue,
  citecolor=gecodeblue,
  filecolor=gecodeblue,
  menucolor=gecodeblue,
  urlcolor=gecodeblue,
  linktocpage=true,
  bookmarksnumbered=true,
  pdftitle={View-based Propagator Derivation},
  pdfauthor={Christian Schulte and Guido Tack},
  plainpages=false,
  pdfpagelabels,
  a4paper]{hyperref}

\newcommand{\httpurl}[1]{\href{http://#1}{\texttt{#1}}}

\usepackage[neverdecrease]{paralist}

\usepackage{dcolumn}

\usepackage{listings}
\usepackage{graphics}
\usepackage{graphicx}
\usepackage{xspace}

\newcommand{\ignore}[1]{}

\newcommand{\pparagraph}[1]{\paragraph{\normalfont\textbf{#1.}}}

\def\CPP{\leavevmode\textrm{\hbox{C\hskip
-0.1ex\raise 0.5ex\hbox{\tiny ++}}}}

\newcommand{\setc}[2]{\ensuremath{\left\{#1\;\middle\vert\;#2 \right\}}}
\newcommand{\tuple}[1]{\ensuremath{\left\langle #1\right\rangle}}

\newcommand{\fun}[2]{\ensuremath{#1\rightarrow #2}}
\newcommand{\restrict}[2]{\ensuremath{{#1}_{|#2}}}
\newcommand{\DEFEQ}{\ensuremath{=}}

\newcommand{\ZZ}{\ensuremath{\mathbb Z}}
\newcommand{\RR}{\ensuremath{\mathbb R}}

\newcommand{\Power}[1]{\ensuremath{\mathcal{P}(#1)}}

\newcommand{\mimpl}{\Longrightarrow}

\newcommand{\mequiv}{\Leftrightarrow}
\newcommand{\lequiv}{\leftrightarrow}

\newcommand{\xor}{\oplus}

\newcommand{\constraintname}[1]{\emph{#1}\xspace}
\newcommand{\alldiff}{\constraintname{all-different}}

\newcommand{\regular}{\constraintname{regular}}
\newcommand{\same}{\constraintname{same}}
\newcommand{\reify}[2]{\left(#1\right)\lequiv #2}

\newcommand{\SetName}[1]{\ensuremath{\mathsf{#1}}\xspace}
\newcommand{\Asn}{\SetName{Asn}}
\newcommand{\Dom}{\SetName{Dom}}
\newcommand{\Con}{\SetName{Con}}
\newcommand{\Prop}{\SetName{Prop}}
\newcommand{\Val}{\ensuremath{V}}
\newcommand{\Var}{\ensuremath{X}}
\newcommand{\ToConstraint}[1]{\llbracket #1\rrbracket}

\newcommand{\sol}{\ensuremath{\operatorname{sol}}}
\newcommand{\vars}{\ensuremath{\operatorname{vars}}}

\newcommand{\stronger}{\subseteq}
\newcommand{\strstronger}{\subset}

\newcommand{\FamUp}[1]{\operatorname{dom}(#1)}
\newcommand{\FamUpR}{\operatorname{dom}_{\RR}}
\newcommand{\FamDown}[1]{\operatorname{con}(#1)}
\newcommand{\boundsd}{\ensuremath{\operatorname{bounds}(D)}\xspace}
\newcommand{\boundsz}{\ensuremath{\operatorname{bounds}(\mathbb{Z})}\xspace}
\newcommand{\boundsr}{\ensuremath{\operatorname{bounds}(\mathbb{R})}\xspace}
\newcommand{\Boundsd}{\ensuremath{\operatorname{Bounds}(D)}\xspace}
\newcommand{\Boundsz}{\ensuremath{\operatorname{Bounds}(\mathbb{Z})}\xspace}
\newcommand{\Boundsr}{\ensuremath{\operatorname{Bounds}(\mathbb{R})}\xspace}
\newcommand{\bnd}{\ensuremath{\operatorname{hull}}}
\newcommand{\bndR}{\ensuremath{\operatorname{hull}_{\RR}}}

\newcommand{\view}{\ensuremath{\varphi}}
\newcommand{\viewA}{\ensuremath{\view_\Asn}}
\newcommand{\viewInv}{\ensuremath{\view^{-}}}
\newcommand{\viewP}[1]{\ensuremath{\widehat{\view}(#1)}}
\newcommand{\viewPPrime}[1]{\ensuremath{\widehat{\view'}(#1)}}

\newcommand{\range}[2]{\left[#1\;..\;#2\right]}

\newcommand{\seq}[1]{\left\langle#1\right\rangle}
\newcommand{\seqc}[3]{\left\langle#1\right\rangle_{#2}^{#3}}
\newcommand{\rseqc}[3]{\seqc{\range{{#1}_{i}}{{#2}_{i}}}{i=1}{#3}}
\newcommand{\rseqcOff}[4]{\seqc{\range{{#1}_{i}+{#4}}{{#2}_{i}+{#4}}}{i=1}{#3}}
\newcommand{\ranges}[1]{\operatorname{ranges}(#1)}

\newcommand{\iinter}[2]{\operatorname{iinter}(#1,#2)}

\newcommand{\iminus}[2]{\operatorname{iminus}(#1,#2)}
\newcommand{\ioffset}[2]{\operatorname{ioffset}(#1,#2)}
\newcommand{\seqset}[1]{\operatorname{set}(#1)}
\newcommand{\iterset}[1]{\operatorname{set}(#1)}
\newcommand{\idone}[1]{\mathit{#1}.\method{done}()}
\newcommand{\inext}[1]{\mathit{#1}.\method{next}()}

\newcommand{\imin}[1]{\mathit{#1}.\method{min}()}
\newcommand{\imax}[1]{\mathit{#1}.\method{max}()}

\newcommand{\xgetdom}[1]{#1.\method{getdom}()}
\newcommand{\xsetdom}[2]{#1.\method{setdom}(#2)}
\newcommand{\xadjdom}[2]{#1.\method{adjdom}(#2)}
\newcommand{\xexcdom}[2]{#1.\method{excdom}(#2)}
\newcommand{\getdom}{\method{getdom}}
\newcommand{\setdom}{\method{setdom}}
\newcommand{\xgetglb}[1]{#1.\method{glb}()}
\newcommand{\xgetlub}[1]{#1.\method{lub}()}
\newcommand{\xadjglb}[2]{#1.\method{adjglb}(#2)}
\newcommand{\xadjlub}[2]{#1.\method{adjlub}(#2)}
\newcommand{\glb}[1]{\operatorname{glb}(#1)}
\newcommand{\lub}[1]{\operatorname{lub}(#1)}

{\vspace{-0.3em}\begin{displaymath}\begin{array}{l@{\;:=\;}l}}%
{\end{array}\end{displaymath}\vspace{-0.6em}}
{\vspace{-0.3em}
\begin{displaymath}\begin{array}{l@{\;:=\;}l@{\qquad}l@{\;:=\;}l}}
{\end{array}\end{displaymath}\vspace{-0.6em}}

\newcommand{\elb}{\ensuremath{\mathsf{lbc}}}
\newcommand{\eub}{\ensuremath{\mathsf{ubc}}}

\newcommand{\method}[1]{\texttt{#1}}

\newcommand{\est}{\ensuremath{\operatorname{est}}}
\newcommand{\lst}{\ensuremath{\operatorname{lst}}}
\newcommand{\ect}{\ensuremath{\operatorname{ect}}}
\newcommand{\lct}{\ensuremath{\operatorname{lct}}}

\title{View-based Propagator Derivation}

\author{Christian Schulte\\
{\small ICT, KTH - Royal Institute of Technology, Sweden, \texttt{cschulte@kth.se}}
\and
Guido Tack\\
{\small Programming Systems Lab, Saarland University, Germany, \texttt{tack@ps.uni-sb.de}}
}
\date{}

\begin{document}

\maketitle

\begin{abstract}
  When implementing a propagator for a constraint, one must
  decide about variants: When implementing $\min$, should one
  also implement $\max$?  Should one implement linear constraints
  both with unit and non-unit coefficients?  Constraint variants
  are ubiquitous: implementing them requires considerable (if not
  prohibitive) effort and decreases maintainability, but will
  deliver better performance than resorting to constraint
  decomposition.
  
  This paper shows how to use views to derive \emph{perfect}
  propagator variants.  A model for views and derived propagators
  is introduced. Derived propagators are proved to be indeed
  perfect in that they inherit essential properties such as
  correctness and domain and bounds consistency. Techniques for
  systematically deriving propagators such as transformation,
  generalization, specialization, and type conversion are
  developed. The paper introduces an implementation architecture
  for views that is independent of the underlying constraint
  programming system.  A detailed evaluation of views implemented
  in Gecode shows that derived propagators are efficient and that
  views often incur no overhead.  Without views, Gecode would
  either require $180\,000$ rather than $40\,000$ lines of
  propagator code, or would lack many efficient propagator
  variants.  Compared to $8\,000$ lines of code for views, the
  reduction in code for propagators yields a $1750\%$ return on
  investment.
\end{abstract}

\section{Introduction}
\label{sec:intro}

When implementing a propagator for a constraint, one typically
must also decide whether to implement some of its variants.  When
implementing a propagator for the constraint
$\max\{x_1,\dots,x_n\}=y$, should one also implement
$\min\{x_1,\dots,x_n\}=y$? The latter can be implemented using
the former as $\max\{-x_1,\dots,-x_n\}=-y$.  When implementing a
propagator for the linear equation $\sum_{i=1}^n a_i x_i=k$ for
integer variables $x_i$ and integers $a_i$ and $k$, should one
also implement the special case $\sum_{i=1}^n x_i=k$ for better
performance?  When implementing a propagator for the reified
linear equation $\reify{\sum_{i=1}^n x_i=c}{b}$, should one also
implement $\reify{\sum_{i=1}^n x_i\neq c}{b}$? These two
constraints only differ by the sign of $b$, as the latter is
equivalent to $\reify{\sum_{i=1}^n x_i=c}{\lnot b}$.

The two straightforward approaches for implementing constraint
variants are to either implement dedicated propagators for the
variants, or to decompose. In the last example, for instance, the
reified constraint could be decomposed into two propagators, one
for $\reify{\sum_{i=1}^n x_i=c}{b'}$, and one for $b\lequiv\lnot
b'$, introducing an additional variable $b'$.

Implementing the variants inflates code and documentation and is
error prone. Given the potential code explosion, one may be able
to only implement some variants (say, $\min$ and $\max$). Other
variants important for performance (say, ternary $\min$ and
$\max$)
may be infeasible due to excessive programming and
maintenance effort. Decomposing, on the other hand, massively
increases memory consumption and runtime.

This paper introduces a third approach: \emph{deriving}
propagators from already existing propagators using \emph{views}.
This approach combines the efficiency of dedicated propagator
implementations with the simplicity and effortlessness of
decomposition.

\begin{example}[Deriving a minimum propagator]
  Consider a propagator for the constraint $\max(x,y)=z$. Given
  three additional propagators for $x'=-x$, $y'=-y$, and $z'=-z$, we
  could propagate the constraint $\min(x',y')=z'$ using the
  propagator for $\max(x,y)=z$. Instead, this paper proposes to derive a 
  propagator using views that perform the simple transformations corresponding  
  to the three additional propagators.
  
  Views transform input and output of a propagator. For example,
  a minus view on a variable $x$ transforms the variable domain
  of $x$ by negating each element, passes the transformed domain
  to the propagator, and performs the inverse transformation on
  the domain returned by the propagator. With views, the
  implementation of the maximum propagator can be reused: a
  propagator for the minimum constraint can be derived from a
  propagator for the maximum constraint and a minus view for each
  variable.
\end{example}


This paper contributes an implementation-independent model for
views and derived propagators, techniques for deriving
propagators, concrete implementation techniques, and an
evaluation that shows that views are widely applicable,
drastically reduce programming effort, and yield an efficient
implementation.

More specifically, we identify the properties of views that are
essential for deriving \emph{perfect} propagators. The paper
establishes a formal model that defines a view as a function and
a derived propagator as functional composition of views (mapping
values to values) with a propagator (mapping domains to
domains). This model yields all the desired results:
derived propagators are indeed propagators; they
faithfully implement the intended constraints; domain consistency
carries over to derived propagators; different forms of bounds
consistency over integer variables carry over provided that the
views satisfy additional yet natural properties.

We introduce techniques for deriving propagators that use views
for transformation, generalization, specialization, and type
conversion of propagators. We show how to apply these techniques
for different variable domains using various views and how
views can be used for the derivation of dual scheduling propagators.

We present and evaluate different implementation approaches for views
and derived propagators. An implementation using parametric
polymorphism (such as templates in \CPP) is shown to incur no
or very low overhead. The architecture is orthogonal to the used
constraint programming system and has been fully implemented in
Gecode~\cite{Gecode}. We analyze how successful
the use of derived propagators has been for Gecode.

\pparagraph{Plan of the paper}

\rule{0pt}{1em} 
\autoref{sec:preliminaries} introduces constraints and
propagators.  \autoref{sec:views} establishes views and
propagator derivation.
\autoref{sec:techniques_for_deriving_propagators} presents
propagator derivation techniques. \autoref{sec:implementation}
describes an implementation architecture based on parametric
propagators and range iterators. \autoref{sec:limitations}
discusses limitations of views. The implementation is evaluated
in \autoref{sec:experimental_evaluation}, and
\autoref{sec:conclusion} concludes.

\section{Preliminaries}
\label{sec:preliminaries}

This section introduces constraints, propagators, and propagation
strength.

\pparagraph{Variables, constraints, and domains}
\label{sub:variables_constraints_domains}

Constraint satisfaction problems use a \define{finite set of
  variables} $\Var$ and a \define{finite set of values} $\Val$.
We typically write variables as $x,y,z\in\Var$ and values as
$v,w\in\Val$.

A solution of a constraint satisfaction problem assigns a single
value to each variable. A constraint restricts which assignments
of values to variables are allowed.

\begin{definition}[Assignments and constraints]\label{def:asncon}
  An \define{assignment} $a$ is a function mapping variables to
  values. The set of all assignments is
  $\Asn\DEFEQ\fun{\Var}{\Val}$.  A \define{constraint} $c$ is a
  set of assignments,
  $c\in\Con\DEFEQ\Power{\Asn}=\Power{\fun{\Var}{\Val}}$ (we write
  $\Power{S}$ for the power set of $S$). Any assignment $a\in c$
  is a \define{solution} of $c$.
\end{definition}

Constraints are defined on assignments as total functions on all
variables. For a typical constraint $c$, only a subset $\vars(c)$
of the variables is \emph{significant}; the constraint is the
full relation for all $x\notin\vars(c)$.  Constraints are either
written as sets of assignments (for example,
$\setc{a\in\Asn}{a(x)<a(y)}$) or as expressions with the usual
meaning, using the notation $\ToConstraint{\cdot}$ (for example,
$\ToConstraint{x<y}$).

\begin{example}[Sum constraint]
  Let $\Var=\{x,y,z\}$ and $\Val=\{1,2,3,4\}$. The constraint
  $\ToConstraint{x=y+z}$ corresponds to the following set of
  assignments:
\begin{align*}
  \ToConstraint{x=y+z} = \{ &(x\mapsto a, y\mapsto b, z\mapsto c)\ |\ 
                            {a,b,c\in V} \land {a=b+c}\}\\
                       = \{ &(x\mapsto 2, y\mapsto 1, z\mapsto 1),
                             (x\mapsto 3, y\mapsto 1, z\mapsto 2),\\
                            &(x\mapsto 3, y\mapsto 2, z\mapsto 1),
                             (x\mapsto 4, y\mapsto 2, z\mapsto 2)\}
\end{align*}
\end{example}

\ignore{
In addition to constraints, constraint satisfaction problems
define domains of variables.
}

\begin{definition}[Domains]
  A \define{domain} $d$ is a function mapping variables to sets
  of values, such that $d(x)\subseteq \Val$. The set of all
  domains is $\Dom\DEFEQ\fun{\Var}{\Power{\Val}}$. The set of
  values in $d$ for a particular variable $x$, $d(x)$, is called
  the \define{variable domain} of $x$.  A domain $d$ represents a
  set of assignments, a constraint, defined as
\begin{equation*}
 \FamDown{d}\DEFEQ\setc{a\in\Asn}{\forall x\in\Var: {a(x)\in d(x)}} 
\end{equation*}
An assignment $a\in\FamDown{d}$ is \define{licensed} by $d$.
\end{definition}

Domains thus represent \emph{Cartesian} sets of assignments.  In
this sense, any domain is also a constraint. For a more uniform
representation, we take the liberty to use domains as
constraints. In particular, $a\in d$ (instead of
$a\in\FamDown{d}$) denotes an assignment $a$ licensed by
$d$, and $c\cap d$ denotes $c\cap\FamDown{d}$.

A domain $d$ that maps some variable to the empty value set is
\define{failed}, written $d=\emptyset$, as it represents no valid
assignments ($\FamDown{d}=\emptyset$).  A domain $d$ representing
a single assignment, $\FamDown{d}=\{a\}$, is \define{assigned},
and is written as $d=\{a\}$.

\begin{definition}[Constraint satisfaction problems]
  A \define{constraint satisfaction problem} (CSP) is a pair
  $\tuple{d,C}$ of a domain $d$ and a set of constraints $C$. The
  \define{solutions} of a CSP $\tuple{d,C}$ are the assignments
  licensed by $d$ that satisfy all constraints in $C$, defined as
  $\sol(\tuple{d,C})\DEFEQ\setc{a\in\FamDown{d}}{\forall c\in C:
    {a\in c}}$.
\end{definition}

\pparagraph{Propagators}
\label{sub:propagators}

A propagation-based constraint solver employs \emph{propagators}
to implement constraints. A propagator for a constraint $c$ takes
a domain $d$ as input and removes values from the variable
domains in $d$ that are in conflict with $c$.

A domain $d$ is \define{stronger} than a domain $d'$, written
$d\stronger d'$, if and only if $d(x)\subseteq d'(x)$ for all
$x\in\Var$. A domain $d$ is \define{strictly stronger} than a
domain $d'$, written $d\strstronger d'$, if and only if $d$ is
stronger than $d'$ and $d(x)\subset d'(x)$ for some variable $x$.
The goal of constraint propagation is to prune values from
variable domains, thus inferring stronger domains, without
removing solutions of the constraints.

A propagator is a function $p$ that takes a domain as its
argument and returns a stronger domain, it may only \emph{prune}
assignments. If the original domain is an assigned domain
$\{a\}$, the propagator either accepts ($p(\{a\})=\{a\}$) or
rejects ($p(\{a\})=\emptyset$) it, realizing a \emph{decision
  procedure} for its constraint. The pruning and the decision
procedure must be consistent: if the decision procedure accepts
an assignment, the pruning procedure must never remove this
assignment from any domain. This property is enforced by
requiring propagators to be monotonic.

\begin{definition}[Propagators]
  A \define{propagator} is a function $p\in\fun{\Dom}{\Dom}$ that
  is
  \begin{itemize}
  \item \define{contracting:} $p(d)\stronger d$ for any domain
    $d$;
  \item \define{monotonic:} $p(d')\stronger p(d)$ for any
    domains $d'\stronger d$.
  \end{itemize}
  The set of all propagators is $\Prop$. If a propagator $p$
  returns a \emph{strictly} stronger domain ($p(d)\strstronger
  d$), we say that $p$ \define{prunes the domain} $d$. The
  propagator $p$ \define{induces} the unique constraint $c_p$ defined by
  the set of assignments accepted by $p$:
\begin{align*}
c_p\DEFEQ\setc{a\in\Asn}{p(\{a\})=\{a\}}  
\end{align*}
\end{definition}

Propagators can also be \define{idempotent} ($p(p(d))=p(d)$ for
any domain $d$).  Idempotency is not required to make propagation
sound or complete, but it can make propagation more
efficient~\cite{Schulte:2008:0:Efficient}.  Like idempotency,
monotonicity as defined here is not necessary for soundness or
completeness of a solver~\cite{Schulte:2009:0:Weakly}. Most
definitions and theorems in this paper are independent of whether
propagators are monotonic or not. Non-monotonicity will thus only
be discussed where it is relevant.

\pparagraph{Propagation strength}
\label{sub:propagation_strength}

Each propagator induces a single constraint, but 
different propagators can induce the same constraint,
differing in \emph{strength}. Typical examples are propagators
for the \alldiff constraint that perform naive pruning when
variables are assigned, or establish bounds
consistency~\cite{Puget:98} or domain
consistency~\cite{Regin:94}.

In the literature, propagation strength is usually defined as a
property of a domain in relation to a constraint. For example, a
domain $d$ is \emph{domain-consistent} (also known as generalized
arc-consistent) with respect to a constraint $c$ if $d(x)$ only
contains values that appear in at least one solution of $c$ for
each variable $x$. As this paper is concerned with propagators,
propagation strength is defined with respect to a propagator.

A propagator $p$ is \define{domain-complete} if any domain it
returns is domain-consistent with respect to $c_p$. For any
constraint $c$, there is exactly one domain-complete propagator
for $c$ (as domains form a lattice). It is defined as
$\hat{p}_c(d)\DEFEQ\FamUp{c_p\cap d}$, where $\FamUp{c}$ is the
\emph{domain relaxation} of $c$, the strongest domain that
contains all assignments of $c$,
$\FamUp{c}=\min\setc{d}{c\stronger d}$.

For constraints over integer variables ($\Val\subseteq\ZZ$),
several weaker notions of propagation strength are known. The
most well-known is \emph{bounds consistency}, which in fact can
mean one of four special cases: range, \boundsd, \boundsz, and
\boundsr consistency (as discussed
in~\cite{Choi:2006:0:Finite,Quimper:2006:0:Efficient}).

The first three differ in whether holes are ignored in the
original domain, in the resulting domain, or in both, in that
order.  Holes in a domain are ignored by the function
$\bnd(d)(x)\DEFEQ\range{\min(d(x))}{\max(d(x))}$, which returns
the convex hull of a variable domain $d(x)$ in \ZZ. \Boundsr
consistency only requires solutions to be found in the
real-valued relaxation of the constraint (written $c_{\RR}$), and
is defined using the real-valued convex hull and domain
relaxation (written $\bndR$ and $\FamUpR$).
The different notions of bounds
consistency give rise to the respective definitions of bounds
completeness.

\begin{definition}[Bounds completeness]
  A propagator $p$ is
  \begin{itemize}
  \item range-complete if and only if $p(d)\stronger
    \FamUp{c_p\cap\bnd(d)}$,
  \item \boundsd-complete if and only if $p(d)\stronger
    \bnd(\FamUp{c_p\cap d})$,
  \item \boundsz-complete if and only if
    $p(d)\stronger\bnd(\FamUp{c_p\cap\bnd(d)})$, and
  \item \boundsr-complete if and only if
    $p(d)\stronger\bndR(\FamUpR(c_{p\RR}\cap\bndR(d)))$
  \end{itemize}
  for any domain $d$.
\end{definition}

\section{Views}
\label{sec:views}

This section defines views and proves properties of
view-derived propagators.

\subsection{Views and Derived Propagators}
\label{sub:model}

Given a propagator $p$, a view is represented by two functions,
$\view$ and $\viewInv$, that can be composed with $p$ such that
$\viewInv\circ p\circ\view$ is the desired derived propagator.
The function $\view$ transforms the input domain, and $\viewInv$
applies the inverse transformation to the propagator's output
domain.

\begin{definition}[Variable views and views]
  A \define{variable view} $\view_x\in\fun{\Val}{\Val'}$ for a
  variable $x$ is an injective function mapping values to values.
  The set $\Val'$ may be different from $\Val$, and the
  corresponding sets of assignments, domains, constraints, and
  propagators are called $\Asn'$, $\Dom'$, $\Con'$, and $\Prop'$,
  respectively.
  
  Given a family of variable views $\view_x$ for all $x\in\Var$,
  we lift them point-wise to assignments: $\viewA(a)(x)\DEFEQ
  \view_x(a(x))$.
  A \define{view} $\view\in\fun{\Con}{\Con'}$ is a family of
  variable views, lifted to constraints:
  $\view(c)\DEFEQ\setc{\viewA(a)}{a\in c}$. The \emph{inverse} of
  a view is defined as
  $\viewInv(c)\DEFEQ\setc{a\in\Asn}{\viewA(a)\in c}$.
\end{definition}

\begin{definition}[Derived propagators and constraints]
  Given a propagator $p\in\Prop'$ and a view $\view$, the
  \define{derived propagator} $\viewP{p}\in\Prop$ is defined as
  $\viewP{p}\DEFEQ\viewInv\circ p\circ\view$. Similarly, a
  \define{derived constraint} is defined to be
  $\viewInv(c)\in\Con$ for a given $c\in\Con'$.
\end{definition}

\begin{example}[Scale views]
\label{ex:scale}
  Given a propagator $p$ for the constraint
  $c=\ToConstraint{x=y}$, we want to derive a propagator for
  $c'=\ToConstraint{x=2y}$ using a view $\view$ such that
  $\viewInv(c)=c'$.
  
  Intuitively, the function $\view$ leaves $x$ as it is and
  scales $y$ by $2$, while $\viewInv$ does the inverse
  transformation. We thus define $\view_x(v)=v$ and
  $\view_y(v)=2v$. That clarifies the need for different
  sets $\Val$ and $\Val'$, as $\Val'$ must contain all
  elements of $\Val$ multiplied by $2$.
  
  The derived propagator is $\viewP{p}=\viewInv\circ
  p\circ\view$.  We say that $\viewP{p}$ ``uses a scale view on''
  $y$, meaning that $\view_y$ is the function defined as
  $\view_y(v)=2v$. Similarly, using an identity view on $x$
  amounts to $\view_x$ being the identity function on $\Val$.
  
  Given the assignment $a=(x\mapsto 2,y\mapsto 1)$, we first
  apply $\view$ and get $\view(\{a\})=\{(x\mapsto 2,y\mapsto
  2)\}$. This is accepted by $p$ and returned unchanged, so
  $\viewInv$ transforms it back to $\{a\}$. Another assignment
  $a'=(x\mapsto 1,y\mapsto 2)$ is transformed to
  $\view(\{a'\})=\{(x\mapsto 1,y\mapsto 4)\}$, rejected
  ($p(\view(\{a'\}))=\emptyset$), and the empty domain is mapped
  to the empty domain by $\viewInv$. The propagator $\viewP{p}$
  induces $\viewInv(c)$.
\end{example}

\subsection{Correctness of Derived Propagators}
\label{sub:correctness_of_derived_propagators}

Derived propagators are well-defined and correct: a derived propagator
$\viewP{p}$ is in fact a propagator, and it induces the desired
constraint ($c_{\viewP{p}}=\viewInv(c_p)$). The proofs of these
statements employ the following direct
consequences of the definitions of views:
\begin{enumerate}[P1.]
\item $\view$ and $\viewInv$ are monotonic by construction (as
  $\view$ and $\viewInv$ are defined
  point-wise)\label{viewsaremonotonic}.
\item $\viewInv\circ\view=\mathrm{id}$ (the identity function, as
  $\view$ is injective)\label{viewsidentity}.
\item $|\view(\{a\})|=1$,
  $\view(\emptyset)=\emptyset$\label{viewsonassigned}.
\item For any view $\view$ and domain $d\in\Dom$, we have
  $\view(d)\in\Dom'$, and for any $d'\in\Dom'$, we have 
  $\viewInv(d')\in\Dom$ (as views are
  defined point-wise)\label{viewsreturndomains}.
\end{enumerate}

\begin{proposition}[Correctness]
\label{theorem:derivedprop}
For a propagator $p$ and view $\view$, $\viewP{p}$ is a
propagator.
\end{proposition}

\begin{proof}
  The derived propagator is well-defined because both $\view(d)$
  and $\viewInv(d)$ are domains (see \propref{viewsreturndomains}
  above). We have to show that $\viewInv\circ p\circ\view$ is
  contracting and monotonic.
  
  For contraction, we have $p(\view(d))\stronger\view(d)$ as $p$
  is contracting. From monotonicity of $\viewInv$ (with
  \propref{viewsaremonotonic}), it follows that
  $\viewInv(p(\view(d)))\stronger\viewInv(\view(d))$. As
  $\viewInv\circ\view=\mathrm{id}$ (with
  \propref{viewsidentity}), we have
  $\viewInv(p(\view(d)))\stronger d$, which proves that
  $\viewP{p}$ is contracting.
  
  Monotonicity is shown as follows for all domains $d',d$ with $d'\stronger d$:
\begin{align*}
&&  \view(d')              &\stronger \view(d)
  && \text{($\view$ monotonic, \propref{viewsaremonotonic})}\\
&\mimpl&  p(\view(d'))           &\stronger p(\view(d))
  && \text{($p$ monotonic)}\\
&\mimpl&  \viewInv(p(\view(d'))) &\stronger \viewInv(p(\view(d)))
  && \text{($\viewInv$ monotonic, \propref{viewsaremonotonic})}
\end{align*}
In summary, for any propagator $p$, $\viewP{p}=\viewInv\circ
p\circ\view$ is a propagator.
\end{proof}

Non-monotonic propagators as defined in~\cite{Schulte:2009:0:Weakly} must at least be \emph{weakly} monotonic, which means that $p(\{a\})\stronger p(d)$ for all domains $d$ and assignments $a\in d$. The above proof can be easily adjusted to weakly monotonic propagators by replacing $d'$ with $\{a\}$ and using \propref{viewsonassigned} in the second line of the proof.

\begin{proposition}[Induced constraints]
\label{thm:viewscorrectness}
Let $p$ be a propagator, and let $\view$ be a view. Then
$\viewP{p}$ induces the constraint $\viewInv(c_p)$.
\end{proposition}

\begin{proof}
  As $p$ induces $c_p$, we know $p(\{a\})= c_p\cap\{a\}$ for all
  assignments $a$. With $|\view(\{a\})|=1$
  (\propref{viewsonassigned}), we have $p(\view(\{a\}))=
  c_p\cap\view(\{a\})$.  Furthermore, we know that
  $c_p\cap\view(\{a\})$ is either $\emptyset$ or $\view(\{a\})$.
\begin{itemize}
  \item \textit{Case $\emptyset$:}
   We have $\viewInv(p(\view(\{a\})))=\emptyset=\setc{a'\in\Asn}{a=a'\land\viewA(a)\in c_p}=\viewInv(c_p)\cap\{a\}$.
 \item \textit{Case $\view(\{a\})$:} With
   \propref{viewsidentity}, we have
   $\viewInv\circ\view=\mathrm{id}$ and hence
   $\viewInv(p(\view(\{a\})))=\{a\}$. Furthermore,
   $\viewInv(c_p)\cap\{a\}=\setc{a'\in\Asn}{a=a'\land
     \viewA(a)\in c_p}=\{a\}$.
\end{itemize}
Together, this shows that $\viewInv\circ
p\circ\view(\{a\})=\{a\}\cap\viewInv(c_p)$.
\end{proof}

Another important property is that views preserve contraction: if
a propagator $p$ prunes a domain, the pruning will not be lost
after transformation by $\viewInv$.

\begin{proposition}[Views preserve contraction]
\label{theorem:contraction}
Let $p$ be a propagator, let $\view$ be a view, and let $d$ be a
domain such that $p(\view(d))\subset \view(d)$. Then
$\viewP{p}(d)\subset d$.
\end{proposition}

\begin{proof}
  The definition of $\viewInv(c)$ is
  $\setc{a\in\Asn}{\viewA(a)\in c}$. Hence,
  $|\viewInv(c)|\leq|c|$. Similarly, we know that
  $|\view(c)|=|c|$. From $p(\view(d))\subset \view(d)$, it
  follows that $|p(\view(d))|<|\view(d)|$. Together, this yields
  $|\viewP{p}(d)|<|\view(d)|=|d|$. We have seen in
  \autoref{theorem:derivedprop} that $\viewP{p}(d)\stronger d$,
  so we can conclude that $\viewP{p}(d)\strstronger d$.
\end{proof}

\subsection{Completeness of Derived Propagators}
\label{sub:completeness_of_derived_propagators}

Ideally, a propagator derived from a domain- or bounds-complete
propagator should inherit its completeness. It turns out to not
generally be true for all notions of completeness and all views.
This section first shows how \boundsz completeness is inherited,
and then generalizes this result to the other notions.

The key insight is that \boundsz completeness of propagators
derived using a view $\view$ depends on whether $\view$ and
$\viewInv$ commute with the $\bnd$ operator, as defined below.
\begin{definition}
  A constraint $c$ is a \define{$\view$-constraint} for a view
  $\view$ if and only if for all $a\in c$, there is a $b\in\Asn$
  such that $a=\viewA(b)$. A view $\view$ is
  \define{hull-injective} if and only if
  $\viewInv(\bnd(\FamUp{c}))=\bnd(\FamUp{\viewInv(c)})$ for all
  $\view$-constraints $c$. It is \define{hull-surjective} if
  and only if $\view(\bnd(d))=\bnd(\view(d))$ for all domains
  $d$. It is \define{hull-bijective} if and only if it is
  hull-injective and hull-surjective.
\end{definition}

The proofs rely on the additional fact that views
commute with set intersection.

\begin{lemma}
\label{lem:viewinter}
For any view $\view$, the equation $\viewInv(c_1\cap
c_2)=\viewInv(c_1)\cap\viewInv(c_2)$ holds.
\end{lemma}

\begin{proof}
By definition of $\viewInv$, we have
\begin{equation*}
 \viewInv(c_1\cap c_2)=\setc{a\in\Asn}{\viewA(a)\in c_1\land \viewA(a)\in c_2} 
\end{equation*}
As $\viewA$ is a function, this is equal to
\begin{equation*}
 \setc{a\in\Asn}{\viewA(a)\in c_1}\cap\setc{a\in\Asn}{\viewA(a)\in c_2}=\viewInv(c_1)\cap\viewInv(c_2) 
\end{equation*}
\end{proof}

\begin{theorem}[\Boundsz completeness]
\label{thm:domsyscplt}
Let $p$ be a \boundsz-complete propagator. For any
hull-bijective view $\view$, the propagator $\viewP{p}$ is
\boundsz-complete.
\end{theorem}

\begin{proof}
From \autoref{thm:viewscorrectness}, we know that $\viewP{p}$ induces the constraint $\viewInv(c_p)$. By monotonicity of $\viewInv$ (with \propref{viewsaremonotonic}) and \boundsz completeness of $p$, we know that
\begin{equation*}
\viewInv\circ p\circ\view(d) \stronger \viewInv(\bnd(\FamUp{c_p \cap \bnd(\view(d))}))
\end{equation*}
We now use the fact that both $\viewInv$ and $\view$ commute with $\bnd(\cdot)$ and set intersection:
\begin{align*}
&\mathrel{\phantom{=}}\viewInv(\bnd(\FamUp{c_p \cap \bnd(\view(d))}))\\
&= \viewInv(\bnd(\FamUp{c_p \cap \view(\bnd(d))}))
&& \text{(hull-surjective)}\\
&=\bnd(\FamUp{\viewInv(c_p \cap \view(\bnd(d)))})
&& \text{(hull-injective)}\\
&=\bnd(\FamUp{\viewInv(c_p) \cap \viewInv(\view(\bnd(d)))})
&& \text{(commute with $\cap$)}\\
&=\bnd(\FamUp{\viewInv(c_p) \cap \bnd(d)})
&& \text{(\propref{viewsidentity})}
\end{align*}
The second step uses hull injectivity, so it requires $c_p \cap \view(\bnd(d))$ to be a $\view$-constraint. All assignments in a $\view$-constraint have to be the image of some assignment under $\viewA$. This is the case here, as the intersection with $\view(\bnd(d))$ can only contain such assignments. So in summary, we get
\begin{equation*}
  \viewInv\circ p\circ\view(d) \stronger
  \bnd(\FamUp{\viewInv(c_p) \cap \bnd(d)}
\end{equation*}
which is the definition of $\viewP{p}$ being \boundsz-complete.
\end{proof}

\pparagraph{Stronger notions of completeness}

Similar theorems hold for domain completeness, range and \boundsz completeness. The theorems directly follow from the fact that any view $\view$ is \define{domain-injective}, meaning that $\viewInv(\FamUp{c})=\FamUp{\viewInv(c)}$ for all constraints $c$. We split this statement into the following two lemmas.

\begin{lemma}
\label{thm:famupisprojection}
Given a constraint $c$, let $d=\FamUp{c}$. Then for all $x\in\Var$, we have $v\in d(x) \mequiv \exists a\in c:\;\ a(x)=v$.
\end{lemma}

\begin{proof}
We prove both directions of the equivalence:
\begin{proofiff}
\NECC There must be such an assignment $a$ because otherwise one can construct a strictly stronger $d'\strstronger d$ with $v\notin d'(x)$ such that still $c\stronger d'$.
\SUFF Each domain $d'$ in the intersection $\bigcap\setc{d'\in\Dom}{c\subseteq\FamDown{d'}}$ must contain the value $v\in d'(x)$ as $c\stronger d'$. So for the result of the intersection $d$, $v\in d(x)$.
\end{proofiff}
\end{proof}

\begin{lemma}
\label{lem:viewdom}
Any view $\view$ is domain-injective.
\end{lemma}

\begin{proof}
We have to show that $\viewInv(\FamUp{c})=\FamUp{\viewInv(c)}$ holds for any constraint $c$ and any view $\view$. For clarity, we write the equation including the implicit $\FamDown{\cdot}$ operations: $\viewInv(\FamDown{\FamUp{c}})=\FamDown{\FamUp{\viewInv(c)}}$.
By definition of $\viewInv$ and $\FamDown{\cdot}$, we have 
\begin{align*}
\viewInv(\FamDown{\FamUp{c}})&=\setc{a\in\Asn}{\forall x\in\Var:\ {\viewA(a)(x)\in\FamUp{c}(x)}}\\
&=
\setc{a\in\Asn}{\forall x\in\Var\;\exists b\in c:\;{\viewA(a)(x)=b(x)}} &\text{(\autoref{thm:famupisprojection})}
\end{align*}
As $\viewA$ is an injective function, we can find such a $b$ that is in the range of $\viewA$, if and only if there is also a $b'\in\viewInv(c)$ such that $\viewA(b')=b$. Therefore, we get
\begin{align*}
&\setc{a\in\Asn}{\forall x\in\Var\;\exists b'\in\viewInv(c):\ {a(x)=b'(x)}}\\
=&\setc{a\in\Asn}{\forall x\in\Var:\ {a(x)\in\FamUp{\viewInv(c)}(x)}}\\
=&\FamDown{\FamUp{\viewInv(c)}}
\end{align*}
\end{proof}

The following three theorems express under which conditions the different notions of completeness are preserved when deriving propagators. The proofs for these theorems are analogous to the proof of \autoref{thm:domsyscplt}, using \autoref{lem:viewdom}.

\begin{theorem}[\Boundsd completeness]
Let $p$ be a \boundsd-complete propagator. For any hull-injective view $\view$, the propagator $\viewP{p}$ is \boundsd-complete.
\end{theorem}

\begin{theorem}[Range completeness]
Let $p$ be a range-complete propagator. For any hull-surjective view $\view$, the propagator $\viewP{p}$ is range-complete.
\end{theorem}

\begin{theorem}[Domain completeness]
Let $p$ be a domain-complete propagator, and let $\view$ be a view.
Then $\viewP{p}$ is domain-complete. \end{theorem}

A propagator derived from a \boundsz-complete
propagator and a hull-injective but not hull-surjective view is only
\boundsr-complete. This is exactly what we would expect from a
propagator for linear equations, as the next example demonstrates.

\begin{example}[Linear constraints] A propagator for a linear constraint
  $c_{\sum}=\ToConstraint{\sum_{i=1}^n x_i=c}$ and $n$ scale
  views (see \autoref{ex:scale}) yield a propagator for a linear
  constraint with coefficients $c_{\sum
    a}=\ToConstraint{\sum_{i=1}^n a_i x_i=c}$.
  
  The usual propagator for a linear constraint with coefficients
  achieves \boundsr consistency in linear
  time~$O(n)$~\cite{newprop-journal}.  However, it \emph{is}
  \boundsz-complete for unit coefficients.  Hence, the
  above-mentioned property applies: The propagator for $c_{\sum}$
  is \boundsz-complete, scale views are only hull-injective, so
  the derived propagator for $c_{\sum a}$ is \boundsr-complete.
  Implementing the simpler propagator without coefficients
  and deriving the variant with coefficients yields propagators
  with exactly the same runtime complexity and propagation
  strength as manually implemented propagators.
\end{example}

\subsection{Additional Properties of Derived Propagators}
\label{sub:more_properties_of_derived_propagators}

This section discusses how views can be composed, and how derived
propagators behave with respect to idempotency and subsumption.

\pparagraph{View composition}

A derived propagator permits further derivation. Consider a
propagator $p$ and two views $\view,\view'$. Then
$\viewPPrime{\viewP{p}}$ is a perfectly acceptable derived
propagator, and properties like correctness and completeness
carry over transitively. For instance, we can derive a propagator
for $\ToConstraint{x-y=c}$ from a propagator for
$\ToConstraint{x+y=0}$, combining an \emph{offset view}
($\view_y(v)=v+c$) and a \emph{minus view} ($\view'_y(v)=-v$) on
$y$. This yields a propagator for
$\ToConstraint{x+(-(y+c))=0}=\ToConstraint{x-y=c}$.

\pparagraph{Fixed points}

Schulte and Stuckey~\cite{Schulte:2008:0:Efficient} show how to
optimize the scheduling of propagators that are known to be at a
fixed point. Views preserve fixed points of propagators, so the
same optimizations apply to derived propagators.

\begin{proposition}
  Let $p$ be a propagator, let $\view$ be a view, and let $d$ be
  a domain. If $\view(d)$ is a fixed point of $p$, then $d$ is a
  fixed point of $\viewP{p}$.
\end{proposition}

\begin{proof}
  Assume $p(p(\view(d)))=p(\view(d))$. We have to show
  $\viewP{p}(d)=\viewP{p}(\viewP{p}(d))$. With the assumption, we
  can write $\viewP{p}(d)=(\viewInv\circ p\circ p\circ\view)(d)$.
  We know that $\view\circ\viewInv(c)=c$ if $|\viewInv(c)|=|c|$.
  As we first apply $\view$, this is the case here, so we can add
  $\view\circ\viewInv$ in the middle, yielding $(\viewInv\circ
  p\circ(\view\circ\viewInv)\circ p\circ\view)(d)$. With function
  composition being associative, this is equal to
  $\viewP{p}(\viewP{p}(d))$.
\end{proof}

\pparagraph{Subsumption}

A propagator is \emph{subsumed} (also known as entailed) by a
domain $d$ if and only if for all stronger domains $d'\subseteq
d$, $p(d')=d'$. Subsumed propagators cannot do any pruning in the
remaining subtree of the search, and can therefore be removed.
Deciding subsumption is coNP-complete in general, but for many
practically relevant propagators an approximation can be decided
easily (such as when a domain becomes assigned). The following
theorem states that the approximation is also valid for the
derived propagator.

\begin{proposition}
\label{theorem:subsumption}
Let $p$ be a propagator and let $\view$ be a view. The propagator
$\viewP{p}$ is subsumed by a domain $d$ if and only if $p$ is
subsumed by $\view(d)$.
\end{proposition}

\begin{proof}
  With \propref{viewsidentity} we get that $\forall {d'\subseteq
    d}:\ {\viewInv(p(\view(d'))) = d'}$ is equivalent to $\forall
  {d'\subseteq d}:\ {\viewInv(p(\view(d'))) =
    \viewInv(\view(d'))}$. As $\viewInv$ is a function, and
  because it preserves contraction (see
  \autoref{theorem:contraction}), this is equivalent to $\forall
  {d'\subseteq d}:\ {p(\view(d')) = \view(d')}$. This can be
  rewritten to $\forall {d''\subseteq \view(d)}:\ {p(d'') = d''}$
  because all $\view(d')$ are subsets of $\view(d)$.
\end{proof}

\subsection{Related Work}
\label{sub:views:related_work}

While the idea to systematically derive propagators using views
is novel, there are a few related approaches we can point out.
Reusing functionality (like a propagator) by wrapping it in an
adaptor (like a view) is of course a much more general
technique---think of higher-order functions like fold or map in
functional programming languages; or chaining command-line tools
in Unix operating systems using pipes.

\pparagraph{Propagator derivation}

Views that perform arithmetic transformations are related to the
concept of indexicals
(see~\cite{Carlsson:1997:0:An-Open-Ended,Hentenryck:1998:0:Design}).
An indexical is a propagator that prunes a single variable and is
defined in terms of range expressions. In contrast to views,
range expressions can involve multiple variables, but on the
other hand only operate in one direction. For instance, in an
indexical for the constraint $\ToConstraint{x=y+z}$, the range
expression $y+z$ would be used to prune the domain of $x$, but
not for pruning the domains of $y$ or $z$. Views must work in both directions, which is why they are limited in
expressiveness.

Unit propagation in SAT solvers performs propagation for Boolean
clauses, which are disjunctions of \emph{literals}, which in turn
are positive or negated Boolean variables. In implementations
such as MiniSat~\cite{Een:2003:0:An-Extensible}, the Boolean
clause propagator is in fact derived from a simple $n$-ary
disjunction propagator and \emph{literal views} of the variables
that perform negation for the negative literals.

\pparagraph{Constraint composition}

Instead of regarding a view $\view$ as \emph{transforming} a
constraint $c$, one can regard $\view$ as \emph{additional}
constraints, implementing the decomposition. Assuming $\vars(c)=x_1,\dots,x_n$, we use
additional variables $x'_1,\dots,x'_n$. Instead of $c$, we use
$c'=c[x_1/x'_1,\dots,x_n/x'_n]$, which is the same relation
as $c$, but on $x'_1,\dots,x'_n$. Finally, $n$ \emph{view
  constraints} $c_{\view,i}$ link the original variables to the
new variables, each $c_{\view,i}$ being equivalent to the
relation $x'_i=\view_i(x_i)$. The solutions of the decomposition
model, restricted to the $x_1,\dots,x_n$, are exactly the
solutions of the original view-based model.

Every view constraint $c_{\view,i}$ shares exactly one variable
with $c$ and no variable with any other $c_{\view,i}$. Thus, the
constraint graph is
Berge-acyclic~\cite{Beeri:1983:0:On-the-Desirability}, and a
fixed point can be computed by first propagating all the
$c_{\view,i}$, then propagating $c[x_1/x'_1,\dots,x_n/x'n]$, and
then again propagating the $c_{\view,i}$. This is exactly what
$\viewInv\circ p\circ\view$ does. Constraint solvers typically do
not provide any means of specifying the propagator scheduling in
such a fine-grained way (Lagerkvist and Schulte show how to use
propagator groups to achieve
this~\cite{LagerkvistSchulte:CP:2009}). Thus, deriving
propagators using views is also a technique for specifying
perfect propagator scheduling.

On a more historical level, a derived propagator is related to
the notion of \emph{path consistency}. A domain is
path-consistent for a set of constraints, if for any subset
$\{x,y,z\}$ of its variables, $v_1\in d(x)$ and $v_2\in d(y)$
implies that there is a value $v_3\in d(z)$ such that the pair
$(v_1,v_2)$ satisfies all the (binary) constraints between $x$
and $y$, the pair $(v_1,v_3)$ satisfies all the (binary)
constraints between $x$ and $z$, and the pair $(v_3,v_2)$
satisfies all the (binary) constraints between $z$ and
$y$~\cite{Mackworth:1977:0:Consistency}. If $\viewP{p}$ is
domain-complete for $\viewInv(c)$, then it achieves path
consistency for the constraint $c[x_1/x'_1,\dots,x_n/x'_n]$ and
all the $c_{\view,i}$ in the decomposition model.

\section{Propagator Derivation Techniques}
\label{sec:techniques_for_deriving_propagators}

This section introduces techniques for deriving propagators using
views. The techniques capture the transformation, generalization,
specialization, and type conversion of propagators and are shown
to be widely applicable across variable domains and application
areas.

\subsection{Transformation}
\label{sec:transformation}

\pparagraph{Boolean connectives}

For Boolean variables, where $\Val=\{0,1\}$, the only view apart
from identity for Boolean variables captures negation.  A
\emph{negation view} on $x$ defines $\view_x(v)=1-v$ for
$x\in\Var$ and $v\in\Val$. As already noted in
\autoref{sub:views:related_work}, deriving propagators using
Boolean views thus means to propagate using \emph{literals}
rather than variables.

The obvious application of negation views is to derive
propagators for all Boolean connectives from just three
propagators. A negation view for $x$ in $x=y$ yields a propagator
for $\neg x=y$. From disjunction $x\vee y=z$ one can derive
conjunction $x\wedge y=z$ with negation views on $x$, $y$, $z$,
and implication $x\rightarrow y=z$ with a negation view on $x$.
From equivalence $x\leftrightarrow y=z$ one can derive exclusive
or $x\xor y=z$ with a negation view on $z$.

As Boolean constraints are widespread, it pays off to optimize
frequently occurring cases of propagators for Boolean
connectives. One important propagator is for $\bigvee_{i=1}^n
x_i=y$ with arbitrarily many variables. Again, conjunction can be
derived with negation views on the $x_i$ and on $y$. Another
important propagator implements the constraint $\bigvee_{i=1}^n
x_i=1$.  A dedicated propagator for this constraint is essential
as the constraint occurs frequently and can be implemented
efficiently using watched literals, see for example
\cite{Gent:2006:0:Watched}. With views
all implementation work is readily reused for conjunction. This
shows a general advantage of views: effort put into optimizing a
single propagator directly pays off for all other propagators
derived from it.

\pparagraph{Boolean cardinality}

Like the constraint $\bigvee_{i=1}^n x_i=1$, the Boolean
cardinality constraint $\sum_{i=1}^n x_i \geq c$ occurs
frequently and can be implemented efficiently using watched
literals (requiring $c+1$ watched literals, Boolean disjunction
corresponds to the case where $c=1$). But also a propagator for
$\sum_{i=1}^n x_i \leq c$ can be derived using negation views on
the $x_i$ with the following transformation:
$$
\begin{array}{rclcl}
\sum_{i=1}^n x_i \leq c &\iff& -\sum_{i=1}^n x_i \geq -c 
&\iff& n-\sum_{i=1}^n x_i \geq n-c \\
&\iff& \sum_{i=1}^n 1-x_i \geq n-c 
&\iff& \sum_{i=1}^n \neg x_i \geq n-c \\
\end{array}
$$

\pparagraph{Reification}

Many reified constraints (such as $\reify{\sum_{x=1}^n
  x_i=c}{b}$) also exist in a negated version (such as
$\reify{\sum_{x=1}^n x_i\neq c}{b}$). Deriving the negated
version is trivial by using a negation view on the Boolean
control variable $b$. This contrasts nicely with the effort
without views: either the entire code must be duplicated or the
parts that perform checking whether the constraint or its
negation is subsumed must be factored out and combined
differently for the two variants.

\pparagraph{Transformation using set views}

Set constraints deal with variables whose values are finite sets.
Using \emph{complement views} (analogous to Boolean negation, as
sets with their usual operations also form a Boolean algebra) on
$x,y,z$ with a propagator for $x\cap y=z$ yields a propagator for
$x\cup y=z$. A complement view on $y$ yields $x\setminus y=z$.

\pparagraph{Transformation using integer views}

The obvious integer equivalent to negation views for Boolean
variables are \emph{minus views:} a minus view on an integer
variable $x$ is defined as $\view_x(v)=-v$. Minus views help to
derive propagators following simple transformations: for example,
$\min(x,y)=z$ can be derived from $\max(x,y)=z$ by using minus
views for $x$, $y$, and $z$.

Transformations through minus views can improve performance in
subtle ways. Consider a $\boundsz$-complete propagator for
multiplication $x\times y=z$ (for example,
\cite[Section~6.5]{Apt:2003} or
\cite{SchulteStuckey:TOPLAS:2005}). Propagation depends on
whether zero is still included in the domains of $x$, $y$, or
$z$. Testing for inclusion of zero each time the propagator is
executed is inefficient and leads to a convoluted implementation.
Instead, one would like to rewrite the propagator to special
variants where $x$, $y$, and $z$ are either strictly positive or
negative. These variants can propagate more efficiently, in
particular because propagation can easily be made idempotent.
Instead of implementing three different propagators ($x,y,z$
strictly positive; only $x$ or $y$ strictly positive; only $z$
strictly positive), a single propagator assuming that all views
are strictly positive is sufficient. The other propagators can be
derived using minus views.

Again, with views it becomes realistic to optimize a single
implementation of a propagator and derive other, equally optimized,
implementations. The effort to implement all required specialized
versions without views is typically unrealistic.

\pparagraph{Scheduling propagators}

An important application area is constraint-based scheduling, see
for example~\cite{BaptisteEa:2001}. Many propagation algorithms
for constraint-based scheduling are based on tasks, where a
task $t$ is characterized by its start time, processing time
(how long does the task take to be executed on a resource), and
end time. Scheduling algorithms are typically expressed in
terms of earliest start time ($\est(t)$), latest start time
($\lst(t)$), earliest completion time ($\ect(t)$), and latest
completion time ($\lct(t)$). 

Another particular aspect of scheduling algorithms is that they
are often required in two, mutually dual, variants.  Let us
consider not-first/not-last propagation as an example.  Assume a
set of tasks $T$ and a task $t\not\in T$ to be scheduled on the
same resource.  Then $t$ cannot be scheduled before the tasks in
$T$ ($t$ is not-first in $T\cup\{t\}$), if $\ect(t) > \lst(T)$
(where $\lst(T)$ is a conservative estimate of the latest start
time of all tasks in $T$). Hence, $\est(t)$ can be adjusted to
leave some room for at least one task from $T$. The dual variant
is that $t$ is not-last: if $\ect(T)>\lst(t)$ (again, $\ect(T)$
estimates the earliest completion time of $T$), then $\lct(t)$
can be adjusted.

\begin{figure}
\begin{center}
\psset{unit=4mm,linewidth=0.2mm}
\begin{pspicture}(-14,1)(14,5)
\psline{->}(-14,2)(14,2)%
\psline(0,1.7)(0,2.3)%
\multirput(1,1.85)(1,0){13}{\psline(0,0)(0,0.15)}%
\multirput(-1,1.85)(-1,0){13}{\psline(0,0)(0,0.15)}%
\psframe[linestyle=dashed](2,3)(11,5)%
\psframe[fillstyle=solid,fillcolor=lightgray](5,3.5)(8,4.5)%
\psline{<->}(2,4)(5,4)\psline{<->}(8,4)(11,4)%
\rput(6.5,4){$t$}
\psframe[linestyle=dashed](-2,3)(-11,5)%
\psframe[fillstyle=solid,fillcolor=lightgray](-5,3.5)(-8,4.5)%
\psline{<->}(-2,4)(-5,4)\psline{<->}(-8,4)(-11,4)%
\rput(-6.5,4){$t'$}
\rput(0,1){\footnotesize$0$}%
\rput(2,1.4){\footnotesize$2$}%
\rput(5,1.4){\footnotesize$5$}%
\rput(8,1.4){\footnotesize$8$}%
\rput(11,1.4){\footnotesize$11$}%
\rput(-2,1.4){\footnotesize$-2$}%
\rput(-5,1.4){\footnotesize$-5$}%
\rput(-8,1.4){\footnotesize$-8$}%
\rput(-11,1.4){\footnotesize$-11$}%
\rput(2,0.2){\footnotesize$\est(t)$}%
\rput(5,0.2){\footnotesize$\ect(t)$}%
\rput(8,0.2){\footnotesize$\lst(t)$}%
\rput(11,0.2){\footnotesize$\lct(t)$}%
\rput(-2,0.2){\footnotesize$\lct(t')$}%
\rput(-5,0.2){\footnotesize$\lst(t')$}%
\rput(-8,0.2){\footnotesize$\ect(t')$}%
\rput(-11,0.2){\footnotesize$\est(t')$}%
\end{pspicture}
\end{center}
\caption{Task $t$ and its dual task $t'$ using a minus view}
\label{fig:task}
\end{figure}
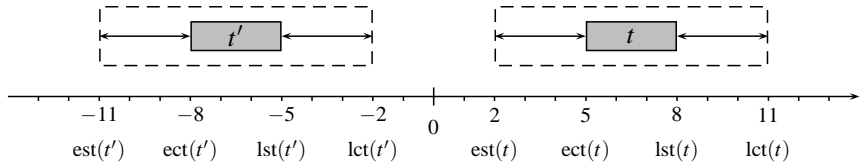

Running the dual variant of a scheduling algorithm on tasks $t\in
T$ is the same as running the original algorithm on the
\emph{dual tasks} $t'\in T'$, which are simply mirrored at the
$0$-origin of the time scale (see \autoref{fig:task}):
$$
\est(t')=-\lct(t) \quad
\ect(t')=-\lst(t) \quad
\lst(t')=-\ect(t) \quad
\lct(t')=-\est(t)
$$
The dual variant of a scheduling propagator can be
automatically derived using a minus view that transforms the time
values. In our example, only a propagator for not-first needs to
be implemented and the propagator for not-last can be derived (or
vice versa). This is in particular beneficial if the algorithms
use sophisticated data structures such as
$\Omega$-trees~\cite{omega-tree}.  Here, also the data structure
needs to be implemented only once and the dual data structure for
the dual propagator is derived.

\subsection{Generalization}
\label{sec:generalization}

Common views for integer variables capture linear transformations
of the integer values: an \emph{offset view} for $o\in\ZZ$ on $x$
is defined as $\view_x(v)=v+o$, and a \emph{scale view} for
$a\in\ZZ$ on $x$ is defined as $\view_x(v)=av$.

Offset and scale views are useful for generalizing propagators.
Generalization has two key advantages: simplicity and efficiency.
A more specialized propagator is often simpler to implement (and
simpler to implement correctly) than a generalized version. The
specialized version can save memory and runtime during
execution.

We can devise an efficient propagation algorithm for a linear
equality constraint $\sum_{i=1}^n x_i=c$ for the common case that
the linear equation has only unit coefficients. The more general
case $\sum_{i=1}^n a_i x_i=c$ can be derived by using scale views
for $a_i$ on $x_i$ (the same technique of course applies to
linear inequalities and disequality rather than equality).
Similarly, a propagator for $\alldiff(x_1,\dots,x_n)$ can be
generalized to $\alldiff(c_1+x_1,\dots,c_n+x_n)$ by using offset
views for $c_i\in\ZZ$ on $x_i$. Likewise, from a propagator for
the element constraint $a[x]=y$ for integers $a_1,\dots,a_n$ and
integer variables $x$ and $y$, we can derive the generalized
version $a[x+o]=y$ with an offset view, where $o\in\ZZ$ provides
a useful offset for the index variable $x$.

These generalizations can be applied to domain- as well as
bounds-complete propagators. While most Boolean propagators are
domain-complete, bounds completeness plays an important role for
integer propagators.
\autoref{sub:completeness_of_derived_propagators} shows that,
given appropriate hull-surjective and/or hull-injective views,
the different notions of bounds consistency are preserved when
deriving propagators.

The views for integer variables presented in this section have
the following properties: minus and offset views are
hull-bijective, whereas a scale view for $a\in\ZZ$ on $x$ is
always hull-injective and only hull-surjective if $a=1$ or $a=-1$
(in which cases it coincides with the identity view or a minus
view, respectively).

\subsection{Specialization}
\label{sec:specialization}

We employ \emph{constant views} to specialize propagators. A
constant view behaves like an assigned variable. In practice,
specialization has two advantages. Fewer variables require less
memory. And specialized propagators can be compiled to more
efficient code, if the constants are known at compile time.

Examples for specialization are
\begin{itemize}
\item a propagator for binary linear inequality $x+y\leq c$
  derived from a propagator for $x+y+z\leq c$ by using a constant
  $0$ for $z$;
\item a reified propagator for $(x=c)\lequiv b$ from
  $(x=y)\lequiv b$ and a constant $c$ for $y$;
\item propagators for the counting constraints
  $|\setc{i}{x_i=c}|=z$ and $|\setc{i}{x_i=y}|=c$ from a
  propagator for $|\setc{i}{x_i=y}|=z$;
\item a propagator for set disjointness from a propagator for $x
  \cap y = z$ and a constant empty set for $z$; and many more.
\end{itemize}

We have to straightforwardly extend the model for constant views.
Propagators may now be defined with respect to a superset of the
variables, $\Var'\supseteq\Var$. A constant view for the value
$k$ on a variable $z\in\Var'\setminus\Var$ translates between the
two sets of variables:
$$
  \view(c) = \setc{a[k/z]}{a\in c}\qquad
  \viewInv(c) = \setc{\restrict{a}{\Var}}{a\in c}
$$
Here, $a[k/z]$ means augmenting the assignment $a$ so that it
maps $z$ to $k$, and $\restrict{a}{\Var}$ is the functional
restriction of $a$ to the set $\Var$.

It is important that this definition preserves failure. If a
propagator returns a failed domain $d$ that maps $z$ to the empty
set, then $\viewInv(d)$ is the empty set, too (recall that this
is really $\viewInv(\FamDown{d})$, and $\FamDown{d}=\emptyset$ if
$d(z)=\emptyset$).

\subsection{Type Conversion}
\label{sec:channeling}

A type conversion view lets propagators for one type of variable
work with a different type, by translating the underlying
representation. Our model already accommodates for this, as a
view $\view_x$ maps elements between different sets $\Val$ and
$\Val'$.

\pparagraph{Integer views}

Boolean variables are essentially integer variables restricted to
the values $\{0,1\}$. Constraint programming systems
may choose a more efficient implementation for Boolean variables and
hence the variable types for integer and Boolean variables
differ. By wrapping an efficient Boolean variable in an
\emph{integer view}, all integer propagators can be directly
reused with Boolean variables. This can save substantial effort:
for example, an implementation of the \regular-constraint for
Boolean variables can be derived which is actually useful in
practice~\cite{LagerkvistPesant:BPPC:2008}.

\pparagraph{Singleton set views}

A \emph{singleton set view} on an integer variable $x$, defined
as $\view_x(v)=\{v\}$, presents an integer variable as a
set variable. Many constraints involve both
integer and set variables, and some of them can be expressed with
singleton set views. A simple constraint is $x\in y$, where $x$
is an integer variable and $y$ a set variable. Singleton set
views derive it as $\{x\}\subseteq y$. This extends to
$\{x\}\diamond y$ for all other set relations $\diamond$.

Singleton set views can also be used to derive pure integer
constraints from set propagators. For example, the constraint
$\same(x_1,\dots,x_n,y_1,\dots,y_m)$ with integer variables
$x_i,y_i$ states that the variables $x_i$ take the same values as
the variables $y_i$. With singleton set views, $\bigcup_{i=1}^n
\{x_i\} = \bigcup_{j=1}^m \{y_j\}$ implements this constraint
(albeit with weaker propagation than the algorithm presented
in~\cite{same}).

\pparagraph{Set bounds and complete set domain variables}

Most systems approximate set variable domains as set intervals
defined by lower and upper
bounds~\cite{Puget:1992:0:PECOS:,Gervet:1994:0:Conjunto:}.
However, \cite{Hawkins:2005:0:Solving} introduces a
representation for the complete domains of set variables, using
ROBDDs. Type conversion views can translate between set interval
and ROBDD-based implementations. We can derive a propagator on
ROBDD-based variables from a set interval propagator, and thus
reuse set interval propagators for which no efficient ROBDD
representation exists.

\subsection{Applicability and Return on Investment}

\begin{table}
\caption{Number of implemented vs.\ derived propagators}
\label{tab:applicability} 
  \footnotesize
  \centering
  \begin{tabular}{|l||D{.}{.}{0}|D{.}{.}{0}|D{.}{.}{2}|}
  \hline
  \emph{Variable type} & \multicolumn{1}{c|}{\emph{Implemented}}    & \multicolumn{1}{c|}{\emph{Derived}} & \multicolumn{1}{c|}{\emph{Ratio}}\\
  \hline\hline
Integer & 77 & 400 & 5.19 \\
Boolean & 28 & 91 & 3.25 \\
Set     & 28 & 122 & 4.36 \\
  \hline
  \emph{Overall} & 133 & 613 & 4.61\\
  \hline
  \end{tabular}
\end{table}

To get an overview of how applicable the presented techniques for
propagator derivation are, let us consider the use of views in
Gecode (version 3.1.0). \autoref{tab:applicability} shows the
number of propagator implementations and the number of
propagators derived from the implementations. On average, every
propagator implementation results in 4.6 derived propagators.
Propagator implementations in Gecode account for more than
$40\,000$ lines of code and documentation. As a rough estimate,
deriving propagators using views thus saves around $140\,000$
lines of code and documentation to be written, tested, and
maintained.  On the other hand, the views mentioned in this
section are implemented in less than $8\,000$ lines of code,
yielding a $1750\%$ return on investment.

\section{Implementation}
\label{sec:implementation}

This section presents an implementation architecture for views
and derived propagators, based on making propagators
\emph{parametric}. Deriving a propagator then means
\emph{instantiating} a parametric propagator with views. The
presented architecture is an orthogonal layer of abstraction on
top of any solver implementation.

\subsection{Views}

The model introduced views as functions that
transform the input and output of a propagator, which maps
domains to domains. In an object-oriented implementation of this
model, a propagator is no longer a function, but an object with a
\method{propagate} method that \emph{accesses} and
\emph{modifies} a domain through the methods of
variable objects. Such an object-oriented model is used for
example by ILOG Solver~\cite{Puget:1994:0:A-CPP-Implementation}
and Choco~\cite{Laburthe:2000:0:CHOCO:}, and is the basis of most
of the current propagation-based constraint solvers.

\begin{figure}[t]
\begin{lstlisting}[language=C++]
  class IntVar {
  private: int _min, _max;
  public:  int min(void) { return _min; }
           int max(void) { return _max; }
           void adjmin(int n) { if (n > _min) _min = n; }
           void adjmax(int n) { if (n < _max) _max = n; }
  };

  class OffsetView {
  protected: IntVar* x; int o;
  public:    OffsetView(IntVar* x0, int o0) : x(x0), o(o0) {}
             int min(void) { return x->min()+o; }
             int max(void) { return x->max()+o; }
             void adjmin(int n) { x->adjmin(n-o); }
             void adjmax(int n) { x->adjmax(n-o); }
  };
\end{lstlisting}
\vspace{-1em}
\caption{Integer variable and offset view}
\label{fig:offsetcpp}
\end{figure}

\autoref{fig:offsetcpp} shows \CPP\ classes for a simple integer
variable (just representing bounds information) and a
corresponding offset view. The view has the same interface as the
variable, so that it can be used in its place. It contains a
pointer to the underlying integer variable and \emph{delegates}
all the operations, performing the necessary transformations.

\subsection{Deriving Propagators}

In order to derive a propagator using view objects like the
above, we use \emph{parametricity}, a mechanism provided by the
implementation language that supports the instantiation of the
same code (the propagator) with different parameters (the views).

\begin{figure}[t]
\begin{lstlisting}[language=C++]
  template<class VX, class VY>
  class Eq : public Propagator {
  protected: VX* x; VY* y;
  public:    Eq(VX* x0, VY* y0) : x(x0), y(y0) {}
             virtual void propagate(void) {
               x->adjmin(y->min()); x->adjmax(y->max());
               y->adjmin(x->min()); y->adjmax(x->max());    
             }
  };
\end{lstlisting}
\vspace{-1em}
\caption{Parametric equality propagator}
\label{fig:eqcpp}
\end{figure}

\autoref{fig:eqcpp} shows a simple equality propagator. The
propagator is based on \CPP\ templates, it is \emph{parametric}
over the types of the two views it uses and can be
\emph{instantiated} with any view that provides the necessary
operations. This type of parametricity is called \emph{parametric
  polymorphism}, and is available in other programming languages
for example in the form of Java
generics~\cite{Gosling:2005:0:JavaTM} or Standard~ML
functors~\cite{Milner:1997:0:The-Definition}.

Given two pointers to integer variables \lstinline!x! and
\lstinline!y!, the propagator can be instantiated to implement
$\ToConstraint{x=y}$ as follows (using the \lstinline!IntVar! class from
\autoref{fig:offsetcpp}):
\begin{quote}
\lstinline!new Eq<IntVar,IntVar>(x,y);!
\end{quote}
\begin{samepage}
The following instantiation yields a propagator for $\ToConstraint{x=y+2}$: 
\begin{quote}
\lstinline!new Eq<IntVar,OffsetView>(x,new OffsetView(y,2));!
\end{quote}
\end{samepage}

\pparagraph{Events}

Most constraint solvers schedule the execution of propagators
according to \emph{events} (see for example
\cite{SchulteCarlsson:CPH:2006}). For example, a propagator $p$
for $\ToConstraint{x<y}$ can only prune the domain (and thus
should only be executed) if either the lower bound of $x$ or the
upper bound of $y$ changes, written $\elb(x)$ and $\eub(y)$. We
say that $p$ \emph{subscribes} to the \emph{event set}
$\{\elb(x),\eub(y)\}$.

Now assume that $p'$ is derived from $p$ using minus views on $x$
and $y$, thus implementing $x>y$. Obviously, $p'$ should
subscribe to the dual event set, $\{\eub(x),\elb(y)\}$. In the
implementation, views provide all the operations needed for event
handling (such as subscription) and perform the necessary
transformations of event sets.

\subsection{Parametricity}

Independent of the concrete implementation, views form an orthogonal
layer of abstraction on top of any propagation-based constraint
solver. As long as the implementation language provides some kind of
parametricity, and variable domains are accessed through some form of
variable objects, propagators can be derived using views.

In addition to parametric polymorphism, two other forms of
parametricity exist, \emph{functional} parametricity and \emph{dynamic
binding}. Functional parametricity means that in languages such as
Standard~ML~\cite{Milner:1997:0:The-Definition} or
Haskell~\cite{Peyton-Jones:2003:0:Haskell}, a higher-order function is
parametric over its arguments. Dynamic binding is typically coupled
with inheritance in object-oriented languages (virtual function calls
in \CPP, method calls in Java). Even in languages that lack direct 
support for parametricity, parametric behavior can often be achieved 
using other mechanisms, such as macros or function pointers in C.

\pparagraph{Choice of parametricity}
In \CPP, parametric polymorphism and dynamic binding have advantages
and disadvantages as it comes to deriving propagators.

Templates are compiled by \emph{monomorphization:} the code is
replicated and specialized for each instance. The compiler can
generate optimized code for each instance, for example by inlining the
transformations that a view implements.

Achieving high efficiency in \CPP{} with templates sacrifices
some expressiveness. Instantiation can \emph{only} happen at
compile-time.  Hence, either \CPP{} must be used for modeling, or
all potentially required propagator variants must be instantiated
explicitly. The \emph{choice} which propagator to use can however
be made at runtime: for linear equations, for instance, if all
coefficients are units, the optimized original propagator can be
posted.

For $n$-ary constraints, compile-time instantiation can be a
limitation, as all arrays must be monomorphic (contain only a single
kind of view). For example, one cannot mix scale and minus views in
linear constraints. For some propagators, we can work around this
restriction using more than a single array of views. For example, a
propagator for a linear constraint can employ two arrays of different
view types, one of which may then be instantiated with identity views
and the other with minus views. While this poses a limitation in
principle, our experience from Gecode suggests that there are
only few propagators in practice that suffer from this limitation.

Dynamic binding is more flexible than parametric polymorphism, as
instantiation happens at runtime and arrays can be polymorphic.
This flexibility comes at the cost of reduced efficiency, as the
transformations done by view operations typically cannot be
inlined and optimized, but require additional virtual method
calls.  \autoref{sec:experimental_evaluation} evaluates
empirically how these virtual method calls affect performance.

\pparagraph{Compile-time versus runtime constants}

Some views involve a parameter, such as the coefficient of a
scale view or the constant of a constant view. These parameters
can again be instantiated at compile-time or at runtime. For
instance, one can regard a minus view as a compile-time
specialization of a scale view with coefficient $-1$, and a zero
view may specialize a constant view. With the constants being
known at compile-time, the compiler can apply more aggressive
optimizations.

\subsection{Iterators}
\label{sec:implementation:iterators}

Typical domain operations involve a single integer value, for
instance adjusting the minimum or maximum of an integer variable.
These operations are not efficient if a propagator performs full
domain reasoning on integer variables or deals with set
variables. Therefore, set-valued operations, like updating a
whole integer variable domain to a new set, or excluding a set of
elements from a set variable domain, are important for
efficiency.  Many constraint programming systems provide an
abstract set-datatype for accessing and updating variable
domains, as for example in Choco~\cite{Choco:2008:0:},
ECL$^i$PS$^e$~\cite{Eclipse:2008:0:}, SICStus
Prolog~\cite{SICStus:2008:0:}, and
Mozart~\cite{The-Mozart-Consortium:2006:0:The-Mozart}. ILOG
Solver~\cite{ILOG:2008:0:ILOG} only allows access by iterating
over the values of a variable domain.

This section develops \emph{iterators} as one particular abstract
datatype for set-valued operations on variables and views. There
are two main reasons to discuss iterators in this paper. First,
iterators provide simple, expressive, and efficient set-valued
operations on variables. Second, and more importantly, iterators
transparently perform the transformations needed for set-valued
operations on views, and thus constitute a perfect fit for
deriving propagators.

\pparagraph{Range sequences and range iterators}

A \define{range} $\range{m}{n}$ denotes the set of integers
$\setc{l\in\ZZ}{m\leq l\leq n}$. A \define{range sequence}
$\ranges{S}$ for a finite set of integers $S\subseteq\ZZ$ is the
shortest sequence
$s=\seq{\range{m_1}{n_1},\ldots,\range{m_k}{n_k}}$ such that
$S=\bigcup_{i=1}^k \range{m_i}{n_i}$ and the ranges are ordered
by their smallest elements ($m_i\leq m_{i+1}$ for $1\leq i<k$).
We thus define the set covered by a range sequence as
$\seqset{s}=\bigcup_{i=1}^k \range{m_i}{n_i}$. The above range
sequence is also written as $\seqc{\range{m_i}{n_i}}{i=1}{k}$.
Clearly, the range sequence of a set is unique, none of its
ranges is empty, and $n_i+1<m_{i+1}$ for $1\leq i<k$.

A \define{range iterator} for a range sequence
$s=\rseqc{n}{m}{k}$ is an object that provides iteration over
$s$: each of the $\range{m_i}{n_i}$ can be obtained in sequential
order but only one at a time. A range iterator $r$ provides the
following operations: $\idone{r}$ tests whether all ranges have
been iterated, $\inext{r}$ moves to the next range, and
$\imin{r}$ and $\imax{r}$ return the minimum and maximum value
for the current range. By $\iterset{r}$ we refer to the set
defined by an iterator $r$ (which must coincide with
$\seqset{s}$).

A range iterator naturally hides its implementation. It can
iterate a sequence (for instance an array) directly by position,
but it can just as well traverse a linked list or the leaves of a
balanced tree, or for example iterate over the union of two other
iterators.

Iterators are consumed by iteration. Hence, if the same sequence
needs to be iterated twice, a fresh iterator is needed. If
iteration is cheap, an iterator can support multiple iterations by providing a 
reset operation. Otherwise, a \emph{cache iterator} takes an arbitrary
range iterator as input, iterates it completely, and stores the
obtained ranges in an array. Its operations then use the array.
The cache iterator implements a reset operation, so that the
possibly costly input iterator is used only once, while the cache
iterator can be used as often as needed.

\pparagraph{Iterators for variables}

The two basic set-valued operations on integer variables are
domain access and domain update. For an integer variable $x$, the
operation $\xgetdom{x}$ returns a range iterator for
$\ranges{d(x)}$. The operation $\xsetdom{x}{r}$ updates the
variable domain of $x$ to $\iterset{r}$ given a range iterator
$r$, provided that $\iterset{r}\subseteq d(x)$. The
responsibility for ensuring that $\iterset{r}\subseteq d(x)$ is
left to the programmer.

In order to provide safer and richer operations, we can use
\emph{iterator combinators}. For example, an \emph{intersection
  iterator} $r=\iinter{r_1}{r_2}$ combines two range iterators
$r_1$ and $r_2$ such that
$\iterset{r}=\iterset{r_1}\cap\iterset{r_2}$. Similarly, a
\emph{difference iterator} $r=\iminus{r_1}{r_2}$ yields
$\iterset{r}=\iterset{r_1}\setminus\iterset{r_2}$.

Richer set-valued operations are then effortless. The operation
$\xadjdom{x}{r}$ adjusts the domain $d(x)$ by $\iterset{r}$,
yielding $d(x)\cap\iterset{r}$, whereas $\xexcdom{x}{r}$ excludes
$\iterset{r}$ from $d(x)$, yielding $d(x)\setminus\iterset{r}$:
\begin{align*}
\xadjdom{x}{r} &\DEFEQ \xsetdom{x}{\iinter{\xgetdom{x}}{r}}\\
\xexcdom{x}{r} &\DEFEQ \xsetdom{x}{\iminus{\xgetdom{x}}{r}}
\end{align*}
In contrast to the $\xsetdom{x}{\cdot}$ operation, the richer
set-valued operations are inherently contracting, and thus safer
to use when implementing a propagator.

Iterators also serve as the natural interface for operations on
set variables, which are usually approximated as
set intervals defined by a lower and an upper
bound~\cite{Puget:1992:0:PECOS:,Gervet:1994:0:Conjunto:}:
$$
d(x)=\range{\glb{d(x)}}{\lub{d(x)}}=\setc{s}{\glb{d(x)}\subseteq s,
  s\subseteq \lub{d(x)}}$$

In order to access and update these set bounds, propagators use
set-valued operations based on iterators: $\xgetglb{x}$ returns a
range iterator for $\ranges{\glb{d(x)}}$, $\xgetlub{x}$ returns a
range iterator for $\ranges{\lub{d(x)}}$, $\xadjglb{x}{r}$
updates the domain of $x$ to $[\glb{d(x)} \cup
\iterset{r},\lub{d(x)}]$, and $\xadjlub{x}{r}$ updates the domain
of $x$ to $[\glb{d(x)},\lub{d(x)}\cap\iterset{r}]$.

Iterator combinators provide the operations that set propagators
need: union, intersection, difference, and complement. Many
propagators can thus be implemented directly using iterators and
do not require any temporary data structures for storing
set-valued intermediate results.

All set-valued operations are parametric with respect to the
iterator $r$: any range iterator can be used. As for parametric
propagators, an implementor has to decide on the kind of
parametricity to use. Gecode uses template-based parametric
polymorphism, with the performance benefits due to
monomorphization and consequent code optimization mentioned
previously.

\pparagraph{Advantages}

Range iterators provide essential advantages over an explicit set
representation. First, any range iterator regardless of its
implementation can be used in domain operations. This turns out
to result in simple, efficient, and expressive domain updates.
Second, no costly memory management is required to maintain a
range iterator as it provides access to only one range at a time.
Third, the abstractness of range iterators makes them compatible
with views and derived propagators: the necessary view
transformations can be encapsulated in an iterator, as discussed
below.

\pparagraph{Iterators for views}

As iterators hide their implementation, they are perfectly suited
for implementing the transformations required for set-valued
operations on views.

Set-valued operations for constant integer views are
straightforward. For a constant view $v$ on constant $k$, the
operation $\xgetdom{v}$ returns an iterator for the singleton
range sequence $\seq{\range{k}{k}}$. The operation
$\xsetdom{v}{r}$ just checks whether the range sequence of $r$ is
empty (in order to detect failure).

Set-valued operations for an offset view are provided by an
\emph{offset iterator}. For a range
sequence $r=\rseqc{m}{n}{k}$ and offset $c$,  $\ioffset{r}{c}$ iterates
$\rseqcOff{m}{n}{k}{c}$. An offset view on $x$ with
offset $c$ then implements $\getdom$ as
$\ioffset{\xgetdom{x}}{c}$ and $\setdom(r)$ as
$\xsetdom{x}{\ioffset{r}{-c}}$.

For minus views we just give the range sequence, iteration is
obvious. For a given range sequence $\rseqc{m}{n}{k}$, the
negative sequence is obtained by reversal and sign change as
$\seqc{\range{-n_{k-i+1}}{-m_{k-i+1}}}{i=1}{k}$. The same
iterator for this sequence can be used both for $\setdom$ and
$\getdom$ operations. Note that implementing the iterator is
involved as it changes direction of the range sequence. There are
two different options for changing direction: either the
set-valued operations accept iterators in both directions or a
cache iterator is used to reverse the direction. Gecode uses the
latter and \autoref{sec:eval:derive} evaluates the overhead
introduced by cache iterators.

A scale iterator provides the necessary transformations for scale
views. Assume a scale view on a variable $x$ with a coefficient
$a>0$, and let $\rseqc{m}{n}{k}$ be a range sequence for $d(x)$.
If $a=1$, the scale iterator does not change the range sequence.
Otherwise, the corresponding scaled range sequence is
$\langle\{a\times m_1\}, \{a\times (m_1+1)\}, \ldots, \{a\times
n_1\}, \ldots, \{a\times m_k\}, $ $ \{a\times (m_k+1)\}, \ldots,
\{a\times n_k\}\rangle$. For the other direction, assume we want
to update the domain using a set $S$ through a scale view. Assume
that $\rseqc{m}{n}{k}$ is a range sequence for $S$. Then for
$1\leq i\leq k$ the ranges $\range{\lceil m_i/a\rceil}{\lfloor
  n_i/a\rfloor}$ correspond to the required variable domain for
$x$, however they do not necessarily form a range sequence as the
ranges might be empty, overlapping, or adjacent. Iterating the
range sequence is however simple by skipping empty ranges and
merging overlapping or adjacent ranges. Scale views for a
variable $x$ and a coefficient $a$ in Gecode are
restricted to be strictly positive so as to not change the
direction of the scaled range sequence. A negative coefficient
can be obtained by using a scale view together
with a minus view.

A complement view of a set variable $x$ uses a \emph{complement
  iterator}, which given a range iterator $r$ iterates over
$\overline{\iterset{r}}$.

\section{Limitations}
\label{sec:limitations}

Although views are widely applicable, they are no silver bullet.
This section explores some limitations of the presented model.

\pparagraph{Beyond injective views}

Views are required to be injective, as otherwise
$\viewInv\circ\view$ is no longer the identity function, and
derived propagators would not necessarily be contracting. An
example for this behavior is a view for the absolute value of an
integer variable. Assuming a variable domain $d(x)=\{1\}$, an
absolute value view $\view$ would leave the domain as it is,
$\view(d)(x)=\{1\}$, but the inverse would ``invent'' the
negative value, $\viewInv(\view(d))(x)=\{-1,1\}$. With an adapted
definition of derived propagators, such as
$\viewP{p}(d)=\viewInv(p(\view(d)))\cap d$, non-injective views
could be used -- however, many of the proofs in this paper rely
on injectivity (though some of the theorems possibly still hold
for non-injective views).

\pparagraph{Multi-variable views}

Some multi-variable views that seem interesting for practical
applications do not preserve contraction, for instance a view on
the sum or product of two variables. The reason is that removing
a value through the view would have to result in removing a
\emph{tuple} of values from the domain. As domains can only
represent Cartesian products, this is not possible in general.
Such a view would have two main disadvantages. First, if
propagation of the original constraint is strong but does not
lead to an actual domain pruning through the views, then the
potentially high computational cost for the pruning does not pay
off. A cheaper but weaker, dedicated propagation algorithm or a
different modeling with stronger pruning is then a better choice.
Second, if views do not preserve contraction, then
\autoref{theorem:subsumption} does not hold. That means that a
propagator $p$ cannot easily detect subsumption any longer, as it
would have to detect it for $\viewP{p}$ instead of just for
itself, $p$. Systems such as Gecode that disable subsumed
propagators (as described in \cite{Schulte:2008:0:Efficient})
then lose this potential for optimization.

For contraction-preserving views on multiple variables, all the
theorems still hold. Two such views we could identify are a set
view of Boolean variables $[b_1,\dots,b_n]$, behaving like
$\setc{i}{b_i=1}$; and an integer view of Boolean variables
$[b_1,\dots,b_n]$, where $b_i$ is $1$ if and only if the integer
has value $i$; as well as the inverse views of these two.

\pparagraph{Propagator invariants}

Propagators typically rely on certain invariants of a variable
domain implementation. If idempotency or completeness of a
propagator depend on these invariants, type conversion views lead
to problems, as the actual variable implementation behind the
view may not respect the same invariants.

For example, a propagator for set variables based on the set
interval approximation can assume that adjusting the lower bound
of a variable does not affect its upper bound. If this propagator
is instantiated with a type conversion view for an ROBDD-based
set variable (see \autoref{sec:channeling}), this invariant is
violated: if, for instance, the current domain is
$\{\{1,2\},\{3\}\}$, and $1$ is added to the lower bound, then
$3$ is removed from the upper bound (in addition to $2$ being
added to the lower bound). If a propagator reports that it has
computed a fixed point based on the assumption that the upper
bound cannot have changed, it may actually not be at a fixed
point. This potentially results in incorrect propagation, for
instance if the propagator could detect failure if it were run
again.

\section{Evaluation}
\label{sec:experimental_evaluation}

While \autoref{sec:views} proved that derived propagators are
perfect with respect to the mathematical model, this section
shows that in most cases one can also obtain perfect
implementations of derived propagators, not incurring any
performance penalties compared to dedicated, handwritten
propagators.

\pparagraph{Experimental setup}

\begin{table}
  \caption{Results for Gecode 3.1.0, the baseline for the experiments}
  \label{tab:baseline}
  \footnotesize
  \centering
\begin{tabular}[t]{|l||D{.}{.}{2}|D{.}{.}{0}|D{.}{.}{0}|D{.}{.}{0}|}
\hline%
\emph{Benchmark}
& \multicolumn{1}{c|}{\emph{time (ms)}}& \multicolumn{1}{c|}{\emph{mem. (KByte)}}& \multicolumn{1}{c|}{\emph{failures }}& \multicolumn{1}{c|}{\emph{propagations }}\\
\hline\hline
All-Interval (50)  & 183.21 & 148 & 0 & 6\,685\\
All-Interval (100)  & 3\,904.21 & 516 & 0 & 25\,866\\
Alpha (naive)  & 100.00 & 23 & 7\,435 & 136\,179\\
BIBD (7-3-60)  & 1\,762.85 & 4\,516 & 1\,306 & 921\,686\\
Eq-20  & 1.52 & 14 & 54 & 3\,460\\
Golomb Rulers (Bnd, 10)  & 423.39 & 67 & 8\,890 & 1\,181\,704\\
Golomb Rulers (Dom, 10)  & 607.86 & 419 & 8\,890 & 1\,181\,770\\
Graph Coloring  & 324.46 & 3\,910 & 1\,100 & 125\,264\\
Magic Sequence (Smart, 500)  & 251.50 & 4\,484 & 251 & 84\,302\\
Magic Sequence (GCC, 500)  & 305.15 & 330 & 251 & 3\,908\\
Partition (32)  & 5\,928.04 & 265 & 160\,258 & 12\,107\,504\\
Perfect Square  & 185.54 & 3\,972 & 150 & 305\,391\\
Queens (10)  & 36.88 & 27 & 4\,992 & 43\,448\\
Queens (Dom, 10)  & 103.38 & 99 & 3\,940 & 59\,508\\
Queens (100)  & 1.54 & 235 & 22 & 455\\
Queens (Dom, 100)  & 31.83 & 2\,056 & 8 & 693\\
Sorting (400)  & 1\,400.01 & 151\,413 & 0 & 459\,501\\
\hline
Social Golfers (8-4-9)  & 193.37 & 10\,254 & 32 & 181\,290\\
Social Golfers (5-3-7)  & 1\,199.51 & 2\,117 & 1\,174 & 852\,391\\
Hamming Codes (20-3-32)  & 1\,140.98 & 24\,746 & 2\,296 & 753\,751\\
Steiner Triples (9)  & 120.11 & 901 & 1\,067 & 297\,501\\
Sudoku (Set, 1)  & 3.48 & 83 & 0 & 1\,820\\
Sudoku (Set, 4)  & 7.30 & 148 & 1 & 3\,752\\
Sudoku (Set, 5)  & 55.14 & 514 & 25 & 28\,038\\
\hline
\end{tabular}
\end{table}

The experiments are based on Gecode~3.1.0, compiled using the GNU
\CPP\ compiler gcc 4.3.2, on an Intel Pentium IV at 2.8 GHz running
Linux. Runtimes are the average of 25 runs, with a coefficient of
deviation less than 2.5\% for all benchmarks. All example programs are
available in the Gecode distribution.
\autoref{tab:baseline} shows the figures for the unmodified
Gecode~3.1.0 (pure integer models above, models with integer and set
variables below the horizontal line), and results will be
given relative to these numbers. For example, a runtime of 130\% means
that the example needs 30\% more time, while 50\% means that it is
twice as fast as in Gecode~3.1.0. The column \emph{time} shows the
runtime, \emph{mem.} the peak allocated memory, \emph{failures} the
number of failures during search, and \emph{propagations} the number
of propagator invocations.

As many of the experimental results rely on the optimization
capabilities of the used \CPP{} compiler, we verified that all
experiments yield similar results with the Microsoft Visual
Studio 2008 \CPP{} compiler.

\subsection{Views Versus Decomposition}

\begin{table}
  \caption{Relative performance of decomposition, compared to views}
  \label{tab:comp:viewsdecomp}
  \footnotesize
  \centering
\begin{tabular}[t]{|l||D{.}{.}{2}|D{.}{.}{2}|D{.}{.}{2}|}
\hline%
\emph{Benchmark}
& \multicolumn{1}{c|}{\emph{time \%}}& \multicolumn{1}{c|}{\emph{mem. \%}}& \multicolumn{1}{c|}{\emph{propagations \%}}\\
\hline\hline
Alpha (naive)  & 412.88 & 360.87 & 673.83\\
BIBD (7-3-60)  & 308.80 & 211.94 & 256.12\\
Eq-20  & 590.35 & 700.00 & 704.57\\
Partition (32)  & 135.61 & 113.58 & 136.40\\
Perfect Square  & 114.46 & 109.67 & 104.42\\
Queens (Dom, 10)  & 173.32 & 100.00 & 519.68\\
Queens (Dom, 100)  & 140.60 & 103.11 & 2\,371.86\\
\hline
Social Golfers (8-4-9)  & 335.89 & 234.82 & 160.22\\
Social Golfers (5-3-7)  & 217.28 & 190.69 & 150.58\\
Hamming Codes (20-3-32)  & 113.81 & 104.66 & 99.65\\
Steiner Triples (9)  & 132.79 & 100.00 & 101.76\\
Sudoku (Set, 1)  & 166.18 & 100.00 & 110.38\\
Sudoku (Set, 4)  & 152.82 & 110.81 & 107.06\\
Sudoku (Set, 5)  & 143.63 & 100.00 & 105.47\\
\hline
\end{tabular}
\end{table}

In order to evaluate whether deriving propagators is worth the
effort in the first place, this set of experiments compares
derived propagators with their decompositions, revealing a
significant overhead of the latter.

\autoref{tab:comp:viewsdecomp} shows the results of these
experiments. For \emph{Alpha} and \emph{Eq-20}, linear equations
with coefficients are decomposed. For \emph{Queens~100}, we
replace the special \alldiff{}-with-offsets by its decomposition
into an \alldiff{} propagator and binary equality-with-offset
propagators. In \emph{BIBD} and \emph{Perfect Square}, we
decompose ternary Boolean propagators, implementing $x\land
y\lequiv z$ as $\lnot x\lor\lnot y\lequiv \lnot z$ in
\emph{BIBD}, and $x\lor y\lequiv z$ as $\lnot x\land\lnot
y\lequiv \lnot z$ in \emph{Perfect Square}. In the remaining
examples, we decompose a set intersection into complement and
union propagators.

Some integer examples show a significant overhead of around
six times the runtime and memory when decomposed. The overhead
of most set examples as well as \emph{Perfect Square} is
moderate, partly because no additional variable was introduced if
the model already contained its complement or negation.
As to be expected, decomposition often needs significantly more
propagation steps, but as the additional steps are performed by
cheap propagators (like $x=y+i$ or $x=\lnot y$), the runtime
effect is less drastic. \emph{Queens~100} is an extreme case,
where 23 times the propagation steps only cause 40\% more
runtime. The reason is that the scheduling order does not take
advantage of the fact that the decompositions are Berge-acyclic
as discussed in \autoref{sub:views:related_work}.
\emph{Partition 32} has a single linear equation with
coefficients, several linear equations with unit coefficients,
multiplications, and a single \alldiff. Replacing the linear
equation by its decomposition has little effect on the runtime
(35\% overhead).

\subsection{Impact of Derivation Techniques}
\label{sec:eval:derive}

The techniques presented in
\autoref{sec:techniques_for_deriving_propagators} have different
impacts on the performance of the derived propagators.

\pparagraph{Generalization and specialization}

These techniques can be implemented without any performance
overhead compared to a handwritten propagator. This is not
surprising as the only potential overhead could be that a
function call is not resolved at compile time. For example, a
thorough inspection of the code generated by the GNU \CPP{}
compiler and the Microsoft Visual Studio \CPP{} compiler shows
that they are able to fully inline the operations of offset and
scale views.

\pparagraph{Transformation and type conversion}

These techniques can incur an overhead compared to a dedicated
implementation, as the transformations performed by the views can
sometimes not be removed by compiler optimizations, and type
conversions may be costly if the data structures for the variable
domains differ significantly.

For example, a propagator instantiated with two minus views of
variables $x$ and $y$ may include a comparison, $(-x)<(-y)$. Due
to the invariants guaranteed by views, this is equivalent to
$y<x$, saving two negations. However, the asymmetry in the two's
complement representation of integers prevents the compiler from
performing this optimization. As an experiment to evaluate this
effect, we instantiated an \alldiff propagator with minus views.
The resulting derived propagator of course implements the same
constraint, but incurs the overhead of negation. Similarly, we
replaced the $\max$ propagator in the \emph{Sort} example with a
$\min$ (where the propagator for $\min$ is derived from the
propagator for $\max$) and negated all parameters. According to
the results in \autoref{tab:comp:minusviews}, the overhead is
often negligible, and only exceeds 5\% in examples that use the
domain-complete \alldiff propagator (\emph{Graph Coloring},
\emph{Golomb Rulers Dom} and \emph{Queens Dom}) or predominantly
$\min$ propagators (\emph{Sort}). \emph{Queens Dom~100} does not
show the effect as the runtime is dominated by search. Using
minus views can result in different propagator scheduling. The
\emph{Partition} example shows this behavior, where the increase
in propagation steps results in increased runtime.

\begin{table}
  \caption{Relative performance of minus views}
  \label{tab:comp:minusviews}
  \footnotesize
  \centering
\begin{tabular}[t]{|l||D{.}{.}{2}|D{.}{.}{2}|}
\hline%
\emph{Benchmark}
& \multicolumn{1}{c|}{\emph{time \%}}& \multicolumn{1}{c|}{\emph{prop. \%}}\\
\hline\hline
All-Interval (50)  & 100.00 & 100.00\\
All-Interval (100)  & 100.52 & 100.00\\
Alpha (naive)  & 101.81 & 100.00\\
Golomb Rulers (Bnd, 10)  & 99.36 & 100.01\\
Golomb Rulers (Dom, 10)  & 107.77 & 100.18\\
Graph Coloring  & 107.19 & 99.84\\
\hline
\end{tabular}
\begin{tabular}[t]{|l||D{.}{.}{2}|D{.}{.}{2}|}
\hline%
\emph{Benchmark}
& \multicolumn{1}{c|}{\emph{time \%}}& \multicolumn{1}{c|}{\emph{prop. \%}}\\
\hline\hline
Partition (32)  & 131.83 & 135.95\\
Queens (10)  & 98.32 & 100.00\\
Queens (Dom, 10)  & 107.63 & 100.00\\
Queens (100)  & 97.83 & 100.00\\
Queens (Dom, 100)  & 95.62 & 100.00\\
Sorting (400)  & 105.13 & 100.00\\
\hline
\end{tabular}
\end{table}

It is interesting to note that the domain-complete \alldiff
propagator, when instantiated with minus views, requires a cache
iterator for sequence reversal (as discussed in
\autoref{sec:implementation:iterators}). Surprisingly, the
overhead of minus views is largely independent of the use of
cache iterators which is confirmed in
\autoref{sec:eval:iterator}.

Other transformations are translated optimally, such as turning
$(-x)-(-y)$ into $y-x$. Boolean negation views also lead to
optimal code, as they do not compute $1-x$ for a Boolean variable
$x$, but instead swap the positive and negative operations.

\begin{table}
  \caption{Relative performance of views compared to dedicated set propagators}
  \label{tab:comp:setviews}
  \footnotesize
  \centering
\begin{tabular}[t]{|l||D{.}{.}{2}|}
\hline%
\emph{Benchmark}
& \multicolumn{1}{c|}{\emph{time \%}}\\
\hline\hline
Social Golfers (8-4-9)  & 166.31\\
Social Golfers (5-3-7)  & 148.83\\
Hamming Codes (20-3-32)  & 129.11\\
Steiner Triples (9)  & 127.85\\
\hline
\end{tabular}
\begin{tabular}[t]{|l||D{.}{.}{2}|}
\hline%
\emph{Benchmark}
& \multicolumn{1}{c|}{\emph{time \%}}\\
\hline\hline
Sudoku (Set, 1)  & 167.44\\
Sudoku (Set, 4)  & 151.74\\
Sudoku (Set, 5)  & 142.83\\
\hline
\end{tabular}
\end{table}

Set-valued transformations can result in non-optimal code. For
example, a propagator for ternary intersection, $x=y\cap z$, will
include an inference $\xadjglb{x}{\xgetglb{y}\cap\xgetglb{z}}$.
To derive a propagator for $x=y\cup z$, we instantiate the
intersection propagator with complement views for $x$, $y$, and
$z$, yielding the following inference:
\begin{equation*}
  \xadjglb{\overline{x}}{\xgetglb{\overline{y}}\cap
                         \xgetglb{\overline{z}}}
\end{equation*}
which amounts to computing
\begin{equation*}
  \xadjlub{x}{\overline{\overline{\xgetlub{y}}\cap
                        \overline{\xgetlub{z}}}}
\end{equation*}
It would be more efficient to implement the equivalent
$\xadjlub{x}{\xgetlub{y}\cup\xgetlub{z}}$ because this requires three 
set operations less. Unfortunately, no compiler will find this
equivalence automatically, as it requires knowledge about the
semantics of the set operations. \autoref{tab:comp:setviews}
compares a dedicated propagator for the constraint $x\cap y=z$
with a version using complement views and a propagator for $x\cup
y=z$. The overhead of 27\% to 67\% does not render views useless
for set variables, but it is nevertheless significant.

\subsection{Templates Versus Virtual Methods}

\begin{table}
  \caption{Relative performance of virtual method calls}
  \label{tab:comp:viewsnoinline}
  \footnotesize
  \centering
\begin{tabular}[t]{|l||D{.}{.}{2}|}
\hline%
\emph{Benchmark}
& \multicolumn{1}{c|}{\emph{time \%}}\\
\hline\hline
All-Interval (50)  & 182.63\\
All-Interval (100)  & 113.20\\
Alpha (naive)  & 153.59\\
BIBD (7-3-60)  & 138.50\\
Eq-20  & 211.69\\
Golomb Rulers (Bnd, 10)  & 220.01\\
Golomb Rulers (Dom, 10)  & 170.13\\
Graph Coloring  & 104.29\\
Magic Sequence (Smart, 500)  & 136.58\\
Magic Sequence (GCC, 500)  & 226.64\\
Partition (32)  & 187.89\\
Perfect Square  & 130.64\\
Queens (10)  & 133.81\\
Queens (100)  & 160.79\\
\hline\end{tabular}
\begin{tabular}[t]{|l||D{.}{.}{2}|}
\hline%
\emph{Benchmark}
& \multicolumn{1}{c|}{\emph{time \%}}\\
\hline\hline
Social Golfers (8-4-9)  & 148.37\\
Social Golfers (5-3-7)  & 138.95\\
Hamming Codes (20-3-32)  & 131.37\\
Steiner Triples (9)  & 149.08\\
Sudoku (Set, 1)  & 119.56\\
Sudoku (Set, 4)  & 118.78\\
Sudoku (Set, 5)  & 119.17\\
\hline
\end{tabular}
\end{table}

As suggested in \autoref{sec:implementation}, in \CPP,
compile-time polymorphism using templates is far more efficient
than virtual method calls. To evaluate this, we changed the basic
operations of integer variables to be virtual methods, such that
view operations need one virtual method call. In addition, all
operations that use templates (and can therefore not be made
virtual in \CPP) have been changed so that they cannot be
inlined, to simulate virtual method calls. This is a conservative
approximation of the actual cost of fully virtual views. The
results of these experiments appear in
\autoref{tab:comp:viewsnoinline}. Virtual method calls cause a
runtime overhead between~4\% and~127\% for the integer examples
(left table), and 18\% to~49\% for the set examples (right
table). The runtime overhead for set examples is lower as the
basic operations on set variables are considerably more expensive
than the basic operations on integer variables.

\subsection{Iterators Versus Temporary Data Structures}
\label{sec:eval:iterator}
  
The following experiments show that using range iterators
improves the efficiency of propagators, compared to the use of
explicit set data structures for temporary results.

\begin{table}[t]
  \caption{Relative performance of cache iterators}
  \label{tab:comp:itertemporary}
  \footnotesize
  \centering
\begin{tabular}[t]{|l||D{.}{.}{2}|}
\hline%
\emph{Benchmark}
& \multicolumn{1}{c|}{\emph{time \%}}\\
\hline\hline
All-Interval (50)  & 102.48\\
All-Interval (100)  & 101.17\\
Golomb Rulers (Bnd, 10)  & 100.51\\
Golomb Rulers (Dom, 10)  & 128.98\\
Graph Coloring  & 144.58\\
Magic Sequence (GCC, 500)  & 103.56\\
Queens (Dom, 10)  & 187.36\\
Queens (Dom, 100)  & 155.62\\
\hline\end{tabular}
\begin{tabular}[t]{|l||D{.}{.}{2}|}
\hline%
\emph{Benchmark}
& \multicolumn{1}{c|}{\emph{time \%}}\\
\hline\hline
Social Golfers (8-4-9)  & 522.62\\
Social Golfers (5-3-7)  & 450.15\\
Hamming Codes (20-3-32)  & 297.38\\
Steiner Triples (9)  & 304.97\\
Sudoku (Set, 1)  & 459.85\\
Sudoku (Set, 4)  & 483.27\\
Sudoku (Set, 5)  & 436.92\\
\hline
\end{tabular}
\end{table}

For the experiments, temporary data structures have been emulated
by wrapping all iterators in a cache iterator as described in
\autoref{sec:implementation:iterators}.
\autoref{tab:comp:itertemporary} shows the results. For integer propagators that perform the safe iterator-based domain operations introduced in \autoref{sec:implementation:iterators}, computing with temporary data structures results in 28\% to 87\% overhead (\emph{Golomb Rulers Dom, Graph Coloring, Queens Dom}). For set propagators, which make much more use of iterators than integer propagators, the overhead becomes prohibitive, resulting in up to 4.8 times the runtime. The memory
consumption does not increase, because iterators are not stored,
and only few iterators are active at a time.

\section{Conclusion}
\label{sec:conclusion}

The paper has developed views for deriving propagator variants. Such
variants are ubiquitous, and the paper has shown how to systematically
derive propagators using different types of views, corresponding to
techniques such as transformation, generalization, specialization, and
type conversion.

Based on a formal, implementation independent model of propagators and
views, the paper has identified fundamental properties of views that
result in \emph{perfect} derived propagators. The paper has shown that
a derived propagator inherits correctness and domain completeness from
its original propagator, and bounds completeness given additional
properties of the used views.

The paper has presented an implementation architecture for views based
on \emph{parametricity}. The propagator implementation is kept
parametric over the type of view that is used, so that deriving a
propagator amounts to instantiating a parametric propagator with the
proper views. This implementation architecture is an orthogonal layer
of abstraction that can be implemented on top of any constraint
solver.

An empirical evaluation has shown that views have proven
invaluable for the implementation of Gecode, saving huge amounts
of code to be written and maintained. Furthermore, deriving
propagators using templates in \CPP\ has been shown to yield
competitive (in most cases optimal) performance compared to
dedicated handwritten propagators. The experiments have also
clarified that deriving propagators is vastly superior to
decomposing the constraints into additional variables and simple
propagators.

\pparagraph{Acknowledgments}

We thank Mikael Lagerkvist, Gert Smolka, and Thibaut Feydy for
fruitful discussions.  Christian Schulte has been partially
funded by the Swedish Research Council (VR) under
grant~621-2004-4953.

\end{document}